\documentclass[letterpaper,10pt,twocolumn]{thuair}
\PassOptionsToPackage{comma,numbers,sort,compress}{natbib}
\usepackage[comma,numbers,sort,compress]{natbib}
\usepackage{graphicx} 
\usepackage{amsmath} 
\usepackage{xspace}
\usepackage{appendix}
\usepackage[misc]{ifsym}
\usepackage{xcolor}
\usepackage{listings}
\lstdefinestyle{lfonts}{
    basicstyle = \footnotesize\ttfamily,
    stringstyle = \color{purple},
    keywordstyle = \color{blue!60!black}\bfseries,
    commentstyle = \color{olive}\scshape,
}
\lstdefinestyle{lnumbers}{
    numbers=left,
    numberstyle=\tiny,
    numbersep=1em,
    firstnumber=1,
    stepnumber=1,
}
\lstdefinestyle{llayout}{
    breaklines=true,
    tabsize=2,
    columns=flexible,
}
\lstdefinestyle{lgeometry}{
    xleftmargin=20pt,
    xrightmargin=0pt,
    frame=tb,
    framesep=\fboxsep,
    framexleftmargin=20pt,
}
\lstdefinestyle{lgeneral}{
    style=lfonts,
    style=lnumbers,
    style=llayout,
    style=lgeometry,
}
\lstdefinestyle{python}{
    language={Python},
    style=lgeneral,
}

\newcommand{\sys}{ChainStream\xspace} 
\newcommand{\bench}{NL-Sense\xspace} 
\newcommand{\agenttt}{\textcolor{black}{sensing-program}\xspace} 
\newcommand{\agenttts}{\textcolor{black}{sensing-programs}\xspace} 
\newcommand{\agentfunc}{\textcolor{black}{stream-function}\xspace} 
\newcommand{\agentfuncs}{\textcolor{black}{stream-functions}\xspace} 

\newcommand{\ie}{\textit{i}.\textit{e}.~}
\newcommand{\eg}{\textit{e}.\textit{g}.~}

\graphicspath{{./images/}} 

\title{ChainStream: An LLM-based Framework for Unified Synthetic Sensing}

\author[1,2,*]{Jiacheng Liu}
\author[1,5,\Letter]{Yuanchun Li}
\author[3]{Liangyan Li}
\author[1]{Yi Sun}
\author[1]{Hao Wen}
\author[1]{Xiangyu Li}
\author[4]{Yao Guo}
\author[1,5]{Yunxin Liu}
\affil[1]{Institute of AI Industry Research (AIR), Tsinghua University}
\affil[2]{Beijing Institute of Technology}
\affil[3]{University of Science and Technology Beijing}
\affil[4]{Peking University}
\affil[5]{Shanghai AI Laboratory}
\affil[*]{Work done during an internship at AIR, Tsinghua University}
\correspondingauthor{liyuanchun@air.tsinghua.edu.cn}

\begin{abstract}
Many applications demand context sensing to offer personalized and timely services. Yet, developing sensing programs can be challenging for developers and using them is privacy-concerning for end-users.
In this paper, we propose to use natural language as the unified interface to process personal data and sense user context, which can effectively ease app development and make the data pipeline more transparent.
Our work is inspired by large language models (LLMs) and other generative models, while directly applying them does not solve the problem - letting the model directly process the data cannot handle complex sensing requests and letting the model write the data processing program suffers error-prone code generation. We address the problem with 1) a unified data processing framework that makes context-sensing programs simpler and 2) a feedback-guided query optimizer that makes data query more informative. To evaluate the performance of natural language-based context sensing, we create a benchmark that contains 133 context sensing tasks. Extensive evaluation has shown that our approach is able to automatically solve the context-sensing tasks efficiently and precisely. The code is opensourced at \href{https://github.com/MobileLLM/ChainStream}{https://github.com/MobileLLM/ChainStream}.
\end{abstract}

\begin{document}
\maketitle

\section{Introduction}


The wide adoption of mobile devices and sensors have enabled a great variety of \textit{context-aware applications (apps)}, \ie the apps driven by situational information about the users, devices, and/or environments.
For example, many apps provide location-based services (LBS) such as ride hailing,  restaurant recommendation, weather reporting, etc. Some apps offer health-related services based on the user's health condition, activity and mood inferred from mobile sensors.
Some apps can also analyze the situations of home, building and city based on distributed cameras.
In the future, as more IoT devices and sensors are being deployed and connected, we anticipate that there will be a huge increasing demand to build context-aware applications.

However, developing context-sensing programs is not easy today, and the challenges are two-fold.
First, the diverse sensor types, data formats and fragmented APIs have made it difficult to write context-sensing programs. Developers usually need to study numerous documentations before actually starting to develop the context-aware apps, which essentially slows down the development and innovation in the field.
Second, it has been a long-standing concern for end-users as such context-aware apps heavily rely on sensitive personal data. Permission-based access control systems in mobile systems (\eg Android) have been criticized almost since its beginning due to its coarse granularity and poor understandability. There is still no better alternative till today.


There are many existing approaches for the above two issues.
For example, to ease the development of context-aware apps, several programming frameworks were introduced, offering better encapsulation of the key APIs required for context sensing \cite{li_imwut17_privacystreams,ferreira2015aware,de2014openpds, liu2024chainstream}. The recent advances of sensing AI have also made it much easier to infer high-level semantic information from different kinds of sensor data.
To reduce users' privacy concerns, researchers have proposed new permission mechanisms \cite{li2016peruim,malviya2022right,aditya2016pic}, app analysis tools \cite{arzt2014flowdroid,enck2014taintdroid}, privacy-aware programming frameworks \cite{taintstream2021,ernst2014collaborative} and annotation systems \cite{jin2022peekaboo,raval2016whatyoumark} to improve the transparency and understandability of personal data access and processing behaviors in apps.
Unfortunately, \textit{these approaches have limited practicability for both developers and end-users}, since the new programming frameworks require developers to learn new APIs, and the abstract concepts (data flow, permission descriptions, etc.) extracted by automated privacy tools are still difficult for users to understand.

Inspired by the recent advances of large language models (LLMs) and large multimodal models (LMMs), we propose to \textbf{use natural language as the unified interface of personal data access and processing in context-aware apps}, which can not only make the app development process much easier (developers can directly use natural language to build the context-sensing parts in their apps, without the need to learn new APIs) but also make the data processing pipeline more transparent (end-users can directly read the developers' natural language data queries and make permission-related decisions).
As a result, such natural language defined context sensing abilities can enable end-user programming of context-aware apps and tasks, potentially fostering more sophisticated usage scenarios in the future.

However, although LLMs and LMMs have demonstrated remarkable data processing and reasoning capabilities, enabling natural language-based context sensing is still challenging.
There are two potential ways of using LLMs/LMMs for context sensing, one is directly using the models to extract the required context information from sensor data, and another is to use the models to generate the code for data processing and context sensing.
The former requires powerful end-to-end data understanding abilities of the foundation model and is unsuitable for processing diverse sensing needs and data modalities.
The later is also hindered by the fragmented sensing APIs and data formats, which make the context sensing programs complex and error-prone.
These difficulties create a large gap in converting high-level natural language context-sensing requests to low-level sensing behavior in applications.

\begin{figure}[tp]
  \centering
  \includegraphics[width=1\linewidth]{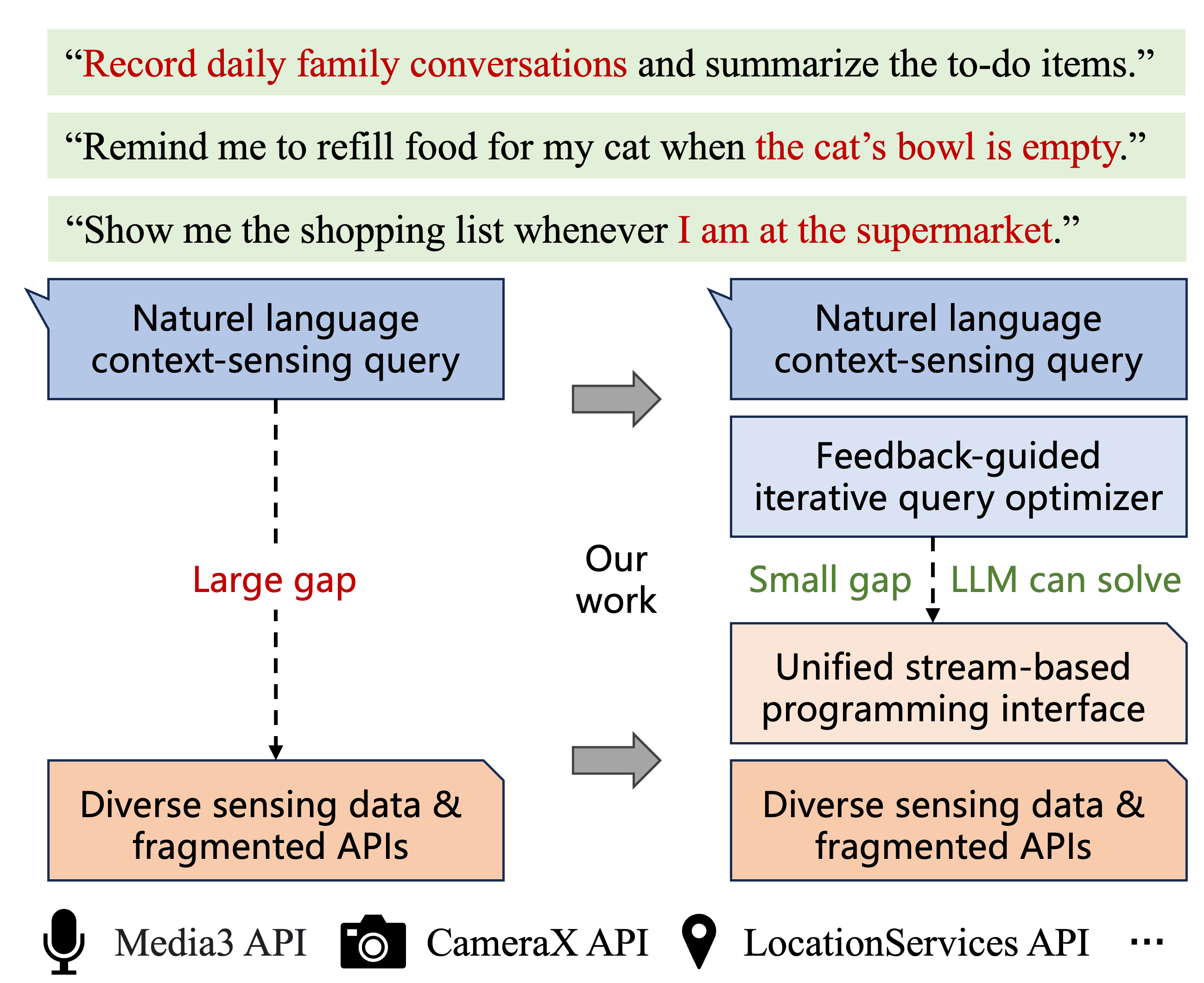}

  \caption{Our basic idea: Enabling natural language-defined context sensing by reducing the gap between the query and the program. }

  \label{fig:bi-directional-approach}
\end{figure}

\textbf{Our approach.}
We take a bi-directional approach to enable precise and flexible natural language based context sensing program generation, as depicted in Figure~\ref{fig:bi-directional-approach}.
One direction is \textbf{to make the sensing program simple and unified} with a stream-style programming framework for sensing and data processing.
Another direction is \textbf{to make the natural language query more informative} by iteratively optimizing the query guided by sandbox feedback.
Doing so can effectively reduce the gap between the natural langauge instructions and the actual context sensing code.

Specifically, in the programming framework, we introduce \texttt{Stream} as the unified abstraction for all kinds of sensing source data and intermediate data. Based on the Stream data abstraction, we design a small unified set of functions for constructing diverse rule-based and model-based stream data operations.
Beneath the programming framework is a runtime system that efficiently manages the data streams and sensing programs with stream flow graphs.
Such a new framework design can effectively reduce the complexity of individual context-sensing programs and the difference between them.
Based on the stream-based programming interface, we further introduce an LLM-based iterative program generator to produce sensing programs from natural language. The generator includes a sandbox-based debugger that provides fine-grained feedback of generated programs and a feedback-guided query optimizer that iteratively refines the LLM queries. After several rounds of refinements, the final programs become more precise and reliable.




\textbf{Benchmark Construction.} 
Since natural language defined context sensing is a relatively new problem, we create a benchmark for evaluating the related approaches. Our benchmark contains 133 tasks, each of which is defined by a natural langauge task description and a ground-truth sensing program. We also include a simulated testbed to examine the generated sensing programs, which contains 16 types of data sources including camera videos, screen recordings, IMU sensor readings, etc.

\textbf{Evaluation.} 
We evaluate our approach on the benchmark in comparison with four strong baselines. The results have shown that our approach is able to achieve highest generation quality, outperforming baselines by about 33\%.

Our work makes the following technical contributions:

\begin{itemize}
    \item To the best of our knowledge, it is the first work on natural language-defined context sensing. We build a new benchmark for this problem.
    \item We introduce an end-to-end system for translating context-sensing requests in natural language to executable sensing programs. The system includes an easy-to-use stream-based programming framework for context sensing and a feedback-guided method to generate context sensing programs.
    \item The evaluation on benchmark has shown the remarkable performance of our approach.
\end{itemize}

\section{Background and Related Work}

\subsection{Context-aware Applications}

Context-aware applications are designed to provide personalized services based on various types of contextual information, collected by both hardware and software sensors. Common contextual information includes the temporal context (e.g., time), spatial context (e.g., position), physical context (e.g., ambient brightness), digital context (e.g., other apps in use), etc.~\cite{incontext_mobile_sensing_nature}

For example, sports apps record users' workout history with time, location, and sensitive health information like heart rate and blood oxygen rate, to help make future plans. Smart home apps manage various connected devices within a home by monitoring physical contexts including temperature, brightness, and humidity. Social apps keep track of various user contexts like the user's location and recent behaviours to provided personalized feeds, recommendations, and deliver targeted advertisements.

\subsection{Developing Context Sensing Programs}

Developing context-aware applications requires considerable efforts due to (1) the diverse and fragmented sensing APIs and (2) users' privacy concerns.
To ease the development process, many programming frameworks have been proposed to provide a unified abstraction of different sensors' APIs, so that the developers can focus more on high-level data flow and context management, instead of low-level implementation details~\cite{li_imwut17_privacystreams,ferreira2015aware,de2014openpds}.
Meanwhile, as context-aware applications rely heavily on sensitive personal data, various privacy-preserving mechanisms have been proposed to address the users' privacy concerns, such as adopting privacy markers, permission mechanisms, app analysis tools, and privacy-aware programming systems~\cite{malviya2022right,arzt2014flowdroid,taintstream2021,jin2022peekaboo,raval2016whatyoumark}. 

Despite the progress made by existing works, the challenges are not completely solved.
New programming frameworks require developers to learn new APIs, which may hinder widespread adoption.
Furthermore, existing privacy protection mechanisms still rely on data flow abstraction and permission descriptions, and access analysis tools only obtain abstract information, which makes it difficult for non-expert users to learn and use.


\subsection{LLM for Sensing and App Development}
Recent advancements in Large Language Models (LLMs) and Large Multimodal Models (LMMs) have enabled a wide range of applications that leverage sensor data for context-aware services, such as health monitoring, activity recognition, and environmental sensing~\cite{foundation_model_as_firmware, li2024personal_llm_agents, llmhealthlearner}. Most existing works directly utilize LLMs/LMMs to analyze sensor data for specific tasks~\cite{dai2024advancing, towardsllmforsensordata, ouyang2024llmsense, xu-etal-2024-penetrative}. These methods have shown promising results in achieving real-world tasks using sensor data. However, they are limited in addressing more complex sensing tasks. For instance, the task of ``recording microphone loudness every two minutes'' requires not only a deep understanding of sensing data but also logic control capabilities, which are better suited to code generation.

An alternative approach is to leverage the code generation capabilities of LLMs to automatically create applications or scripts for sensing-related tasks. Several frameworks have been proposed to handle complex tasks using multi-agent systems~\cite{hong2024metagpt}, self-refinement~\cite{yao2023react}, or in-context learning~\cite{patel-etal-2024-evaluating}. However, these methods fall short in building real-world, context-aware applications due to fragmented sensing APIs and inconsistent data formats. Our approach provides a generalized framework that simplifies the process for LLMs to generate code to accomplish sensing-related tasks effectively.

\subsection{Goal and Challenges}
We aim to introduce a framework where developers and users can access and process personal data using natural language. Doing so can effectively simplify the development of context-aware apps and make the data processing pipeline more transparent for end-users.
However, designing such a framework presents several challenges. First, generating executable code for complex sensing tasks is difficult, even for skilled developers. 
Sensor data is diverse, ranging from physical signals to digital text, and is produced in massive quantities, with streams generated every second. Writing code to handle these tasks efficiently is challenging. 
Second, the fragmented and inconsistent nature of sensor-related APIs leads to extensive documentation, which is difficult for LLMs to comprehend and navigate. 
This complexity makes it challenging for LLMs to effectively work with these APIs, further complicating the development of context-aware applications.






\section{Programming Framework Design}



\begin{figure}[thp]
  \centering
  \includegraphics[width=1\linewidth]{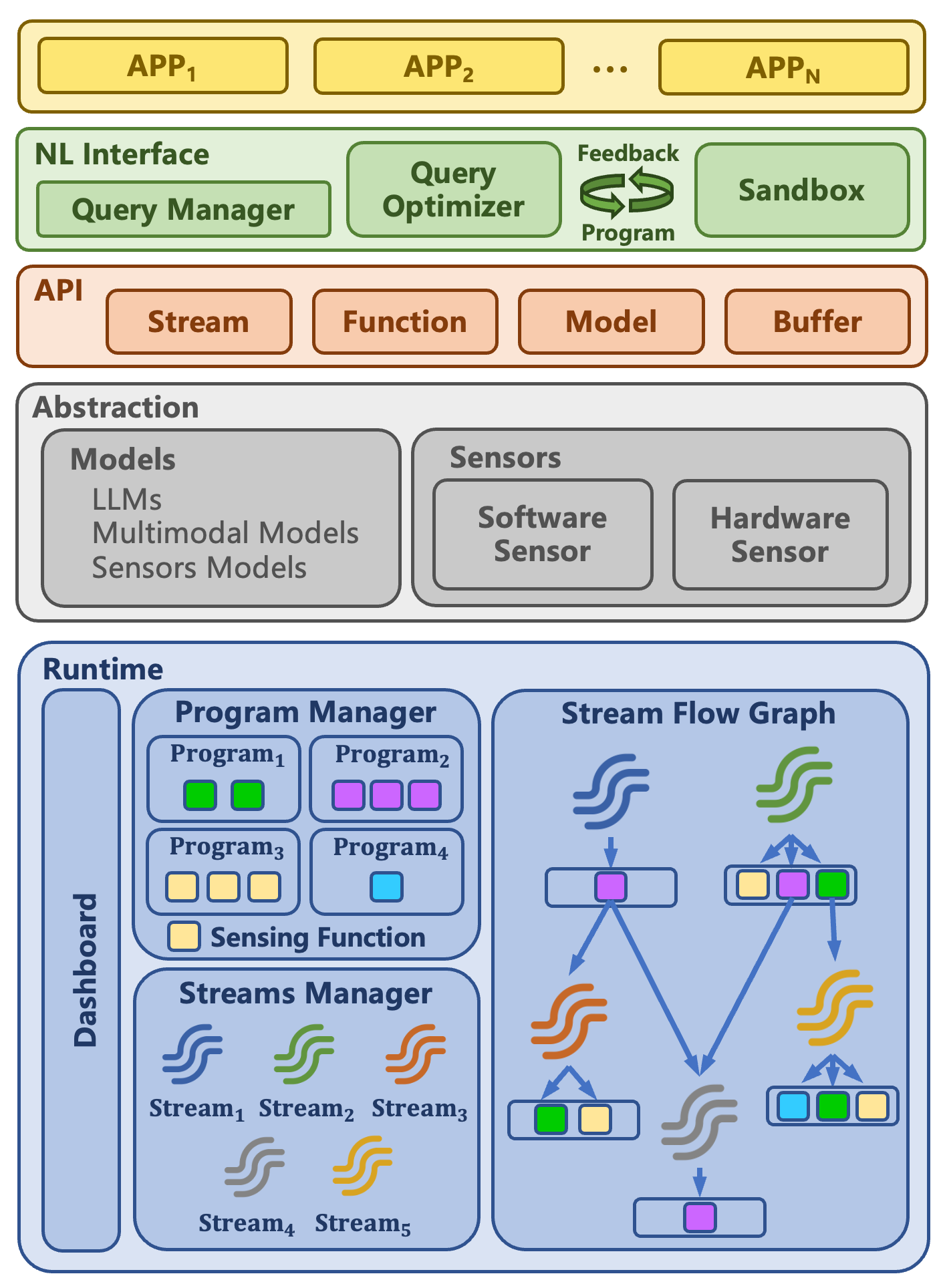}

  \caption{The architecture of \sys programming framework.}

  \label{fig:chainstream_arch}
\end{figure}

Although the existing app development frameworks can already enable experienced developers to write sophisticated context-aware applications, it is still challenging to automatically generate the programs with LLMs.
The purpose of our new programming framework is to address this issue, making context-aware programming easier not only for humans, but also for LLMs.
Although the framework design and implementation mainly involve engineering efforts, it is a key enabler and important basis for further algorithm-related studies.

The design of \sys framework follows two principles: unification and simplification.
Through unification, we aim for different sensing programs to share common patterns, allowing the knowledge gained from one program to be flexibly transferred to others. Simplification involves reducing the number of APIs and statements required to implement each task, making the generated program shorter and less error-prone.

Inspired by the success of stream processing frameworks (\eg Flink, Spark, etc.) in data processing, we introduce a stream-based context-sensing framework in \sys, as shown in Figure~\ref{fig:chainstream_arch}.
Using the stream-style interface offers two important benefits.
First, the stream-based programming model naturally makes the data-processing program more concise, consistent and less fragmented, making it easier to learn for both human and AI programmers.
Second, today's LLMs are typically pretrained with large-scale code corpora, which already contains many programs based on existing popular stream processing framework. Such knowledge can be transferred to our new framework.

The framework mainly consists of three layers: the API layer, the Abstraction layer, and the Runtime layer.
At the API layer, we introduce a set of unified Stream APIs for all kinds of context-sensing tasks. Developers are expected to manipulate the streams with nested custom functions, generate the desired event streams, and trigger actions by listening to the events.
The Abstraction layer offers both a unified model abstraction and a unified data abstraction. The former enables the unified access and management of different foundation models, which can be used to construct various types of AI-based stream transformation functions. The latter enables unified access to and management of sensing data sources, which can be transformed into other streams for more advanced objectives.
Finally, the Runtime layer effectively manages the sensing programs with the Stream Flow Graph (SFG), making the system efficient and transparent.

\subsection{Unified Stream Abstraction}
\label{sec:stream_abstraction}
\subsubsection{Core Concepts: Stream, Function, Program}


\sys adopts the concept of \emph{streams} to abstract all kinds of source data. Each stream can be conceptualized as a dynamic collection of data items that grows infinitely over time. Each data item is a dictionary of key-value pairs with fixed key names and value types. To help developers and AI models to understand, each stream in \sys is associated with a standard \textbf{stream description}. A stream description is a dictionary with three keys: a \texttt{stream\_id} identifier, a natural language stream description, and a dictionary describing the stream item fields, including the type and meaning of each field. 

\sys's computational process draws inspiration from data-flow computing \cite{culler1986dataflow}. \textbf{The functions registered to listen to a stream are referred to as \agentfuncs}. The execution of these \agentfuncs is triggered only when all required operands (\ie required items from different streams) are ready, and the resulting output then flows to the new stream and triggers the next \agentfuncs awaiting this data item.
By applying different types and combinations of \agentfuncs to different input streams, one can achieve different context-sensing objectives and obtain the target event streams. \textbf{We abbreviate the context-sensing programs constructed with \sys as \agenttts}.


Developers using \sys are expected to design \agentfuncs and select the streams to read from and write to based on their sensing goals. The Runtime manages multiple streams globally and employs an event-driven structure to schedule the parallel execution of multiple \agentfuncs as data flows through the system. 

\subsubsection{Abstraction Layer}


Abstractions are used to unify the formats of data and program patterns. The most important aspects include sensor abstraction and model abstraction.

\textbf{Sensor Abstraction.}
Sensing data is the basic material for constructing \agenttts. By nature, different sensors are developed by different providers, using different data formats and APIs. 
To enable the diverse sensing data to be manipulated by our unified API, we encapsulate different sensors within the same Stream abstraction.
Specifically, each abstracted sensor in \sys would produce a stream of items in a predefined format.
Currently, \sys provides 16 built-in sensor abstractions, as shown in Table \ref{tab:raw_sensor_api}. These include common hardware sensors (cameras, microphones, IMU, etc.) and software sensors (\ie continuous stream of data items/events about the user/device/environment) \cite{li2024personal_llm_agents}. 



\begin{table}
  \caption{{Built-in sensor abstractions in \sys.}}

  \label{tab:raw_sensor_api}
  \centering
  \begin{tabular}{ccc}
    \toprule
    Sensor Name & Sensor Type & Data Modality \\
    \midrule
    Camera & Hardware & Image  \\
    Microphone & Hardware & Audio  \\
    Accelerometer & Hardware & Number  \\
    GPS & Hardware & Number  \\
    Gyroscope & Hardware & Number  \\
    Humidity & Hardware & Number  \\
    Light & Hardware & Number  \\
    Pressure & Hardware & Number  \\
    Temperature & Hardware & Number  \\
    Screen Capture & Software & Image \\
    Email & Software & Text \\
    GitHub Events & Software & Text \\
    Daily ArXiv & Software & Text \\
    Daily News & Software & Text \\
    Daily Stocks & Software & Text \\
    Message & Software & Text \\
    \bottomrule
  \end{tabular}
\end{table}


Note that the sensing data may originate from different devices. Therefore, to set up the sensor abstraction, we also develop the network abstraction so that the sensing data streams can be seamlessly centralized on the same device.


\textbf{Model Abstraction.}
An important feature of \sys is to use foundation models to construct the stream processing functions.
However, the models are another source of heterogeneity—foundation models from different developers typically use different APIs and different error handling logic.
To reduce the cognitive load on users of our framework, \sys uses a type-based encapsulation for model selection and use.
The model user only needs to specify the model by the types of input/output data (using \texttt{get\_fm} API), and use the model with the same prompt construction pipeline (using \texttt{build\_prompt} API). Such abstractions also enable flexible and seamless model replacement in the event of model service errors.

\subsubsection{API Design}



    


\begin{table}
  \caption{{Main API descriptions in \sys.}}
  \label{tab:framework_api}
  \resizebox{\columnwidth}{!}{
  \centering
  \begin{tabular}{lll}
    \toprule
     API & Description \\
    \midrule
     \texttt{get\_stream()} & Get a stream by id. \\
       \texttt{create\_stream()} & Create a stream instance. \\
       \texttt{\textbf{Stream}.for\_each()} & Register a function to a stream. \\
       \texttt{\textbf{Stream}.batch()} & Group items into batches. \\
       \texttt{\textbf{Stream}.add\_item()} & Add an item to a stream. \\
       \hline
     \texttt{\textbf{Buffer}.append()} & Add an item to a buffer. \\
      \texttt{\textbf{Buffer}.pop()} & Pop an item from a buffer. \\
      \texttt{\textbf{Buffer}.pop\_all()} & Pop all items from a buffer. \\
      \hline
     \texttt{get\_fm()} & Get a foundation model by input/output types. \\
     \texttt{build\_prompt()} & Create a prompt with multi-modal data. \\
      \texttt{\textbf{Model}.query()} & Query the model with a prompt. \\
    \bottomrule
  \end{tabular}
  }
\end{table}


Table \ref{tab:framework_api} lists the main APIs in the system, including three major categories: \texttt{Stream}, \texttt{Buffer}, and \texttt{Model}. 
When developing a \agenttt, developers need to declare how the program manipulates the streams in the main method.
The \texttt{Stream} API abstracts data streams and provides operations for creating, getting, listening to, writing, and batching streams. The \texttt{Buffer} API is a data structure for caching and managing the intermediate stream data across different streams. 
The \texttt{Model} API encapsulates large language models (LLMs) and other multimodal foundation models, which can be used to construct different intelligent \agentfuncs. 
The \texttt{get\_fm} API is used to obtain a foundation model based on the specified input/output types (\eg text -> text, image+text -> text, etc). The \texttt{build\_prompt} API can assemble a list of data elements with different types into a model prompt, and the \texttt{Model.query} API sends a prompt to the model and returns the model output. 

Compared to other streaming frameworks such as Flink, LangChain \cite{Chase_LangChain_2022} and PrivacyStream \cite{li_imwut17_privacystreams} which often have more than dozens of APIs, our framework aims to reduce the number of APIs to make it easier for an LLM to understand and use. 
However, simplicity does not mean the API is limited. In fact, we can flexibly construct more complex operations (\eg stream joins, aggregations, etc.) and build \agenttt to fulfill diverse sensing tasks. 
We will not provide the detailed definitions and descriptions of each API due to page limits.
The following code snippet illustrates some basic usages of our APIs to achieve more complex tasks.

\begin{lstlisting}[style=python]
model1 = get_fm('image->text')
model2 = get_fm('text->text')

def func1(item: StreamItem):
  prompt = build_prompt('describe the image', item.img)
  description = model1.query(prompt)
  item['desc'] = description
  stream2.add_item(item)
  return item

def func2(item: BatchedItem):
  item_descs = i['desc'] for i in item.batch_items
  prompt = build_prompt('summarize text', item_descs)
  summary = model2.query(prompt)
  item['summary'] = summary
  return item

camera_stream.for_each(func1)
      .batch(by_time='24h')
      .for_each(func2, to_stream=stream3)
\end{lstlisting}
The above code produces two new streams from \texttt{camera\_stream}, including a stream of photo descriptions (\texttt{stream2}) and a stream of daily summaries of captured photos (\texttt{stream3}).

\subsection{Graph-based Runtime Management}



With our stream-based API, each \agenttt in \sys can be viewed as a graph, where the nodes are functions and edges are streams flowing between the functions. The responsibility of the runtime is to manage the programs with a global stream flow graph (SFG for short).
The streams and \agentfuncs use an event-driven model - \agentfuncs are registered in advance to target streams, and are triggered to process new data only when such data appears in the target streams. For managing concurrency in the global computation graph, the system allocates an independent thread for each stream. When a thread has no \agentfunc registered, it remains idle. Otherwise, it operates normally using the event-driven model. 
Different \agentfunc can be flexibly run in parallel (and their model requests can be batched), if they don't have data dependencies in the SFG. Additionally, each stream is allocated a message queue to facilitate inter-thread communication.

To manage the SFG, the \sys runtime includes various system services,
which are responsible for managing, maintaining, and scheduling different resources (processor, memory, model, etc.). 
Additionally, we have various loggers to track basic system information such as stream traffic, program overhead, model usage, and more. Ultimately, this logging information is aggregated into a unified dashboard, supporting analysis, debugging, and adjustments as needed.

The runtime offers an additional benefit of more flexible information sharing and program reuse. In many mobile/edge scenarios, there are often many repetitive sensing processes. The global stream computation graph and atomic function approach in \sys inherently support flexible stream reuse. During \agenttt development, many basic sensing tasks (such as recognizing the current location and behavior of system users) can be accomplished by directly calling existing streams produced by other \agenttt, eliminating the need for redundant coding. 

\section{Iterative Program Generation}

\begin{figure}
  \centering
  \includegraphics[width=1\linewidth]{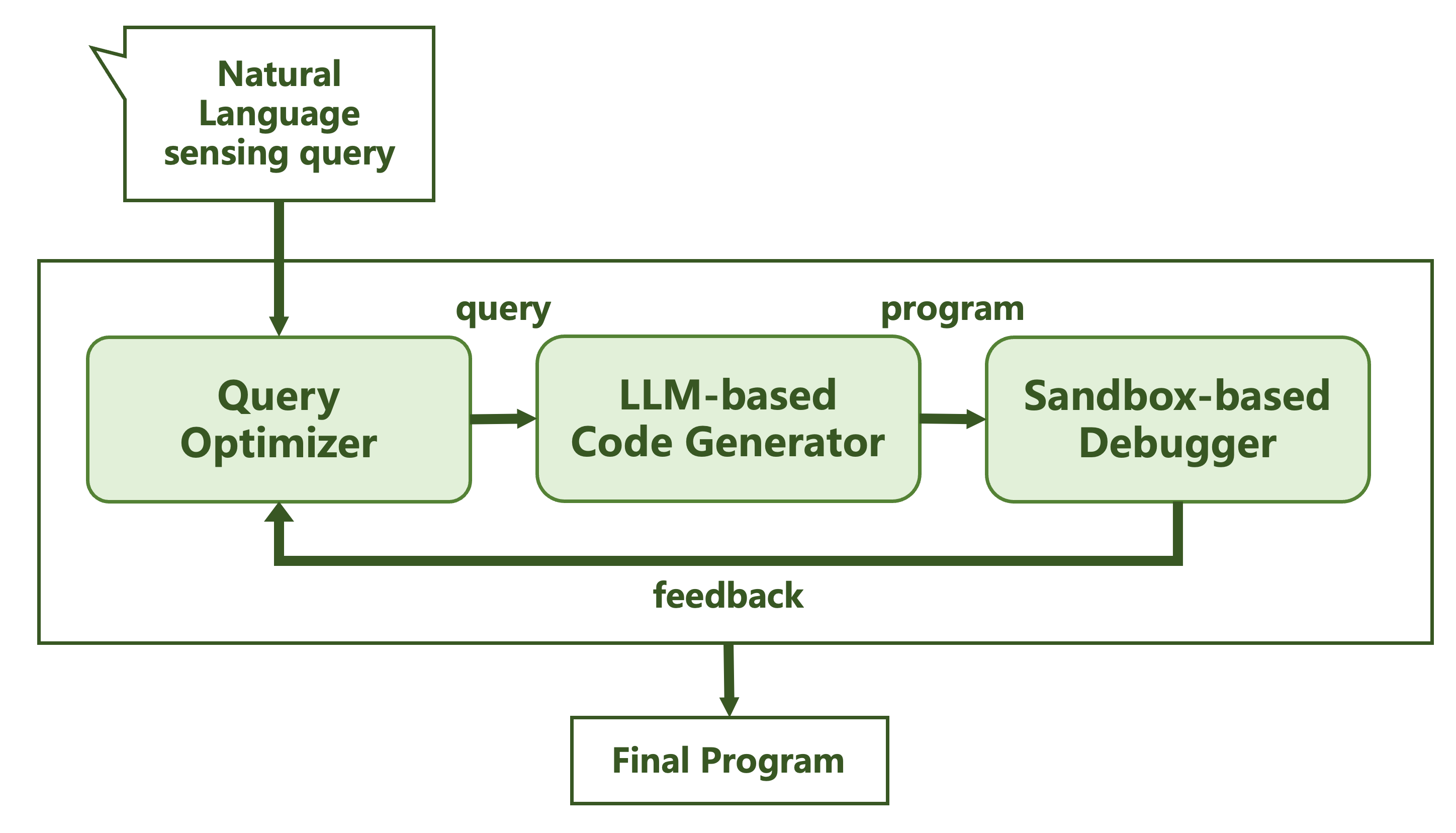}

  \caption{The workflow of Iterative Program Generator in \sys.}

  \label{fig:generator_arch}
\end{figure}



\begin{figure*}
    \centering
    \includegraphics[trim=14pt 212pt 145pt 140pt,clip,width=1\linewidth]{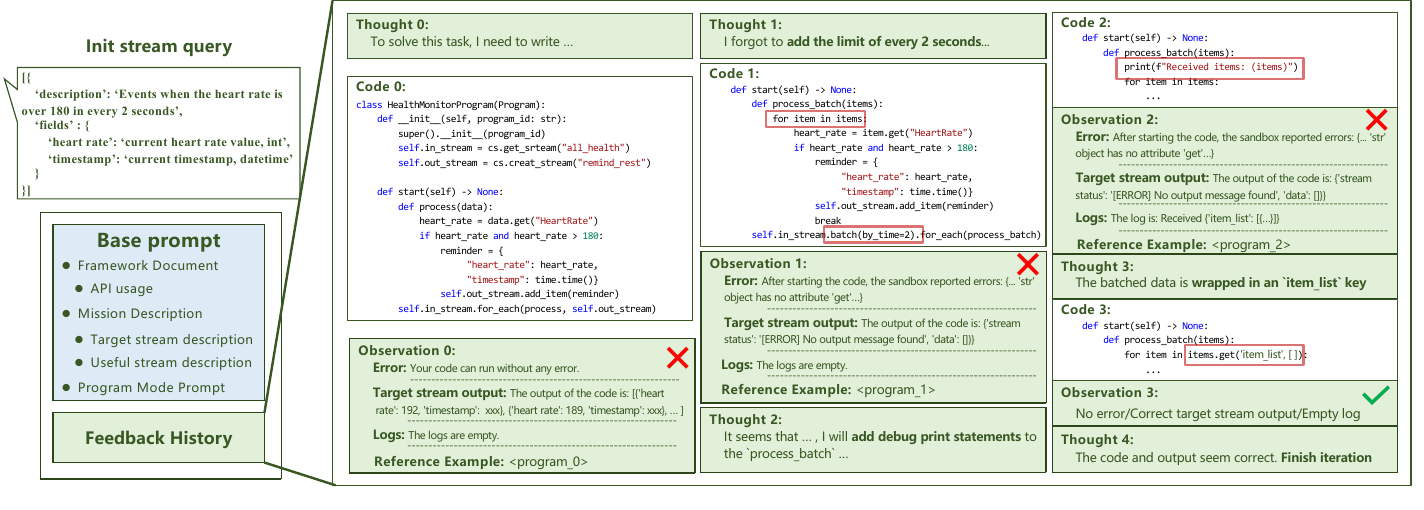}

    \caption{An illustration of the augmented context-sensing query. `Initial stream query' is the original natural language-based sensing query formatted as an expected stream description. The augmented query contains a base prompt and multiple historical sandbox feedbacks.}
    \label{fig:generator_arch_example}
\end{figure*}

Based on the new programming framework, we are closer to achieving our ultimate goal - using natural language to process personal data and sense the context. 
In this section, we introduce the design of \sys Generator, which automatically generates programs based on the \sys programming framework using LLMs.
The generator acts as a programming assistant that continuously interacts with our framework and receives feedback. The generated programs are gradually optimized during the interactions.

The workflow of \sys Generator is shown in Figure \ref{fig:generator_arch}.
The whole process involves three key components, the Query Optimizer, the LLM-based Code Generator, and the Sandbox-based Debugger.

When \sys Generator receives a natural language request for context sensing, \eg ``the microphone loudness value in every 2 minutes'', \sys first passes the query into the Query Optimizer, which will extend the query (by adding the framework document, formatting guidelines, etc.) and return a much more informative one.
The optimized query is then passed to a code-generation LLM (\eg GPT-4) to generate the corresponding sensing program.
Thanks to the unified and concise design of \sys programming framework, complex sensing tasks can be achieved with just a few lines of code.
After a program is generated, it is executed in a sandbox environment. We implement the sandbox environment so that different \sys programs can be interpreted with simulated data. During the simulated program execution, the sandbox records the \agenttt's behavior and generates a report, containing different kinds of program execution feedback.
This feedback can further be used to optimize the program generation query, leading iteratively to better generation results.



Such a feedback-guided program generation process in \sys Generator is analogous to the human programming process. The extended task description is like an interface pending implementation, which includes the expected target (output streams) it needs to return and the input parameters (source streams) it can use, all defined using the structured formats mentioned earlier.
Additionally, due to the in-context learning mechanisms of LLMs, we need to provide the LLMs with a how-to document that introduces the usage of \sys, just as developers need to study the manual before starting to use a new library.
The feedback-guided refinement process is also similar to the debugging process of human developers.

In the program generation workflow, the LLM-based code generator is a minor component that simply involves LLM invocation and result parsing. Therefore, we mainly introduce the other two components including the Sandbox-based Debugger and the Feedback-guided Query Optimizer.

\subsection{Sandbox-based Program Debugger}


The purpose of the debugger is to provide early feedback to the program generator to refine the generation process.
However, debugging stream-processing programs is challenging \cite{banken2018debugging}. The difficulty arises primarily because the stream programs are executed in a lazy, concurrent manner (functions are executed on demand rather than explicitly being called in the control flow). The difficulty is similar in \sys.Once started, the various \agentfuncs within a \agenttt are disaggregated, attached to many system components and distributed across different parts of the system. This distribution causes the error paths to differ from the original locations in the static \agenttt code. Additionally, the system behavior becomes more complex due to this distribution.

To facilitate better stream-based program debugging, we design the \sys Sandbox. The sandbox provides a primary operating environment similar to \sys Runtime, while with simpler control flow transformations, reduced parallelism, and more detailed checking messages. It allows to set up a simulated environment by passing in a \textbf{Env} structure, which defines the available streams, data, and recorded content. A sandbox simulation can be started by specifying a given \textbf{\agenttt} to be analyzed and the environment description (Env) to test the program against.

Specifically, a simulation environment is described with the input streams that can be used by the \agenttt under test and the output streams that the \agenttt is expected to produce.
The input/output streams are described with the standard format defined in Section~\ref{sec:stream_abstraction}.
When testing different \agenttts in the environment, the sandbox simulates the input data streams, feeds them into the programs, records the program behavior, and collects the produced streams.
The recorded runtime behavior and the comparison between the expected output streams and the actually collected streams are the sources of sandbox debugging information.



The lifecycle of a \agenttt in \sys Sandbox includes initialization, starting, running, and stopping. The sandbox captures error information, logs, behavior and output for the target \agenttt at each stage. The error information is detailed down to specific APIs and includes comprehensive error messages. The log is the content printed by the \agenttt during execution. 
The behavior data encompasses all actions of the \agenttt within the \sys system, including timing, content, results, and more. The output data is the output of the target stream. Ultimately, all of this data forms a complete \agenttt behavior report. 
This report is a critical resource for human developers and automated generators to test and evaluate \agenttts. 
As the framework designers, we heavily rely on this report and the simulation sandbox for debugging during the framework development, we will also demonstrate later how important the sandbox report is for the LLM-based programmers. 

\subsection{Feedback-guided Query Refinement}

The Query Optimizer module is responsible for making the original context-sensing query more informative so that LLMs can generate better \agenttts.
It involves two processes: initial refinement and feedback-based incremental refinement.
As shown in Figure~\ref{fig:generator_arch_example}, the initial refinement produces the Base Prompt, and the incremental refinement appends the Feedback History part to the query.

\textbf{The initial refinement} adds necessary background information to the original sensing query, as well as prompts to guide the thinking and output formatting \cite{wei2022cot, yao2023react}.
The base prompt first introduce the usage of \sys, including how to use various modules and the inputs and outputs of each module. Due to context length limitations, we can only use concise language to summarize the main APIs of the system. Fortunately, thanks to the minimalist API design of \sys, we can comprehensively cover all core usages within a short context length. Achieving similar conciseness in other frameworks is challenging. For instance, popular LLM agent frameworks (\eg LangChain, MetaGPT, etc.) typically contain hundreds of core APIs, which are relatively more complex for in-context learning with most LLMs.
To help LLMs understand how to use the APIs to accomplish different sensing goals, we follow the standard prompt engineering practices (such as Chain-of-Thought \cite{wei2022cot}).

The initial refinement process also retrieves the potentially relevant streams and includes the stream descriptions in the base prompt.
The retrieval process is also based on LLMs, which involve providing the desired tasks and all available stream descriptions to the LLM and letting it decide which streams might be useful.
Thanks to the unified and concise abstraction of different streams, the LLMs can precisely (with over 90\% accuracy) identify the most useful input streams for constructing sensing programs.
The prompt produced after the initial refinement can already be used to generate programs with LLMs.

\textbf{The incremental refinement} process further optimizes the program iteratively by dynamically adding organized feedback from the sandbox to the prompt.
Each iteration includes the previous LLM generation output (\emph{Thought} and \emph{code}), and the aggregated feedback information from the sandbox (\emph{Observation}).
The \emph{Thought} component outlines the generator’s reflections on the task and the current state of the solution. The \emph{Code} component consists of the generated programs, which are parsed from the LLM output and executed within our sandbox. 

We now focus on introducing the \emph{Observation} component. 
First, it has a comprehensive error message if the program execution encounters any error. As mentioned before, the sandbox reformulate the program's control flow to make the error reporting more timely and informative, locating the mistakes in code more precisely.
Second, once the program becomes executable, our sandbox will discover its output streams and record the produced stream items. If the produced stream items do not comply with the given output expectation, a message describing the mismatch will also be attached to the feedback.
Third, based on the execution error or stream mismatching information, the feedback also includes a reference example, which is a correct \agenttt (for other context-sensing purpose) that contains the correct usage of the corresponding APIs. The reference selection process is based on random selection after filtering out the irrelevant examples (not using the error-related APIs).

By iteratively incorporating new observation and regenerating the \agenttt, we gradually optimize the generated program. The generation process is terminated either when the generator is satisfied with the current generation (by explicitly returning a `Finish' signal) or after a fixed number of iterations.









\section{Benchmark}
\label{sec:benchmark}

Since natural language-based context sensing is a new problem, there is no existing benchmark for evaluating such systems. Therefore, we create a benchmark (named \bench) that includes a set of manually crafted tasks for context sensing and data processing, as well as the oracle programs (based on \sys API) to solve this task.

\subsection{Tasks and Data}



\textbf{Tasks.}
We collect and summarize the context sensing and data processing demands of various common scenarios in daily life, and then construct the \textbf{Task} set in our benchmark based on these demands. 
Each task mainly consists of four parts: \textbf{target stream description}, \textbf{available stream description}, \textbf{available stream items} and \textbf{oracle code}.
The \textbf{available streams} mimic most of the existing source sensor streams mentioned in Table \ref{tab:raw_sensor_api}, while the \textbf{target streams} represent the functionalities that the generators under evaluation need to achieve.

\begin{figure}
    \centering
    \includegraphics[width=1\linewidth]{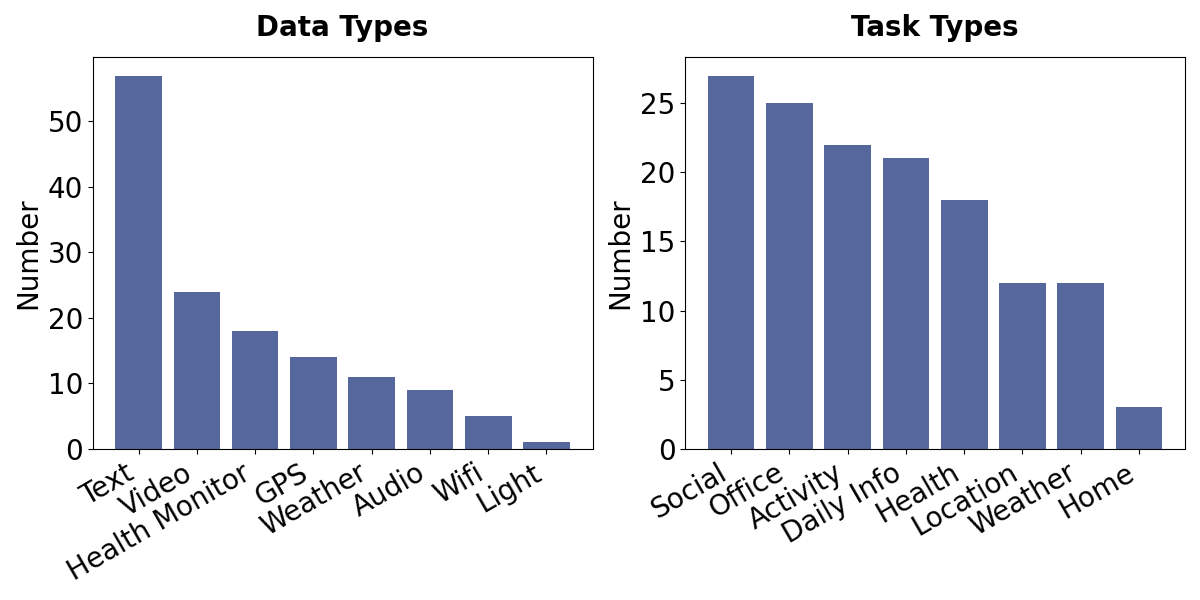}

    \caption{The distribution of task types and data types in our benchmark.}
    \label{fig:task_analysis}
  \end{figure}

Our principles for designing tasks include: 1. Closely aligning with useful context sensing and data processing demands in real life. 2. Covering various scenarios, modalities, difficulty levels, and processing methods as much as possible. 3. Limiting the ambiguity of the tasks, including the randomness of the results and the diversity of the processing methods. Although randomness and diversity are main characteristics of real-world perception and data processing programs, we have to sacrifice some of the flexibility to achieve relative stability in the results. 
Following these principles, we crafted 133 tasks in the benchmark.
The statistical distribution of the tasks is shown in Figure \ref{fig:task_analysis}.
Among them, 87 tasks are relatively simple, focusing primarily on basic stream data processing such as filtering and aggregation, with each task involving only one processing function. 46 tasks are more complex, involving more intricate multi-stream inputs and outputs, with each task involving multiple functions.

\textbf{Stream Simulation.} Our sandbox and runtime can be used to determine whether a generated program is executable. However, determining the extent of correctness of the generated program remains challenging. To do so, we need to test the generated programs by actually \textbf{executing} them with data and \textbf{comparing} the results with the oracle program's output. 
To execute the programs, it is necessary to feed them with meaningful data streams.
This is similar to the common evaluation processes in online judge (OJ) systems and AI-based code generation systems based on test cases, such as HumanEval \cite{HumanEval}, HumanEval-X \cite{zheng2023codegeex} and BigCodeBench \cite{zhuo2024bigcodebench}. However, our benchmark has more challenges including the diversity of modalities, the non-uniqueness and time dynamics of input and output streams, the randomness of LLM-based data transformations, and the logical consistency of results.

\begin{table}
    \centering
    \caption{The datasets used to simulate source streams in our benchmark.}
    \label{tab:raw_dataset_tasks}
    \resizebox{\columnwidth}{!}{
    \begin{tabular}{cccc}
        \toprule
        \textbf{Dataset} & \textbf{Data Type} & \textbf{Description} \\
        \midrule
        Ego4d \cite{ego4d} & Video & First-person visual perception \\
        SPHAR \cite{sphar} & Video & Third-person visual perception \\
        Daily Dialog \cite{dialog} & Audio Text & Audio recordings of conversations \\
        desktop-ui-dataset \cite{desktop_ui} & Image & Screenshots of desktop \\
        Rico \cite{rico} & Image & Screenshots of Android \\
        ArXiv \cite{arxiv} & Text & Scientific papers \\
        Email \cite{emails} & Text & Email messages \\
        GitHub \cite{GitHub_Dataset} & Text & GitHub repositories with details \\
        News \cite{news} & Text & News reports \\
        NUS SMS Corpus \cite{SMS} & Text & SMS messages \\

        Stocks \cite{stock}& Text & Stock market information \\
        Twitter \cite{twitter} & Text & Twitter updates \\

        GPS \cite{world_gpsworld_gps, sea_building_energy}& Sensor & GPS coordinates \\
        Health Monitor \cite{maternal_helth, sleep_dataset} & Sensor & Health data \\
        Sensor readings & Sensor & Light intensity, temperature, humidity etc. \\
        Fitness Track \cite{activities} & Sensor & Activity data \\
        Weather \cite{Weather_Data} & Sensor & Weather data \\
        Wi-Fi \cite{wifi} & Sensor & Wi-Fi information \\

        \bottomrule
    \end{tabular}
    }
\end{table}

We incorporate existing datasets to simulate the sensing data streams. We have summarized and collected various datasets from the real world, as shown in Table \ref{tab:raw_dataset_tasks}. Then we implement different data interface encapsulations for these datasets, which form the raw data streams and make manual adjustments to ensure task validity. These datasets cover multiple modalities, including Image, Audio, Text, and Number. Since LLMs are not suitable for directly processing large amounts of raw numeric sensor data, such as accelerometer and GPS data, we have converted these source streams into streams processed using common models in their respective fields. For example, accelerometer data is converted into basic actions, GPS data into geographical locations, and speech into transcribed text. 

We have not implemented a timing system to simulate the time dynamics of real data streams. Instead, we primarily use the manually adjusted relative order of multi-stream items as the main evaluation criterion. For time-related tasks (such as minute-wise sensing tasks), we modify the interval unit to a scale of seconds in the task description to achieve a relative time scale (for example, a day is longer than an hour, but the exact ratio may not be precise). Creating a perfect time simulation system is a significant challenge that we may attempt to address in the future.

\subsection{Evaluation Method}

\textbf{Oracle Programs.} Our benchmark also includes an oracle program for each task to evaluate the generated program's logical correctness and performance in a comparative manner. Because context sensing tasks are inherently complex and numerous, it is challenging to provide manually labeled outputs. Therefore, the task execution results are examined by comparing with the behavior and output of the oracle program under the same sensing objective.
Specifically, we use the outputs of these standard programs produced by running them in the sandbox as the reference outputs.
We also provide a batch testing interface, which allows evaluating the programs in a parallel, high-throughput manner.





\textbf{Comparison Metrics.}
The quality of the generated sensing program is quantified with several metrics to compare the actual outputs and expected reference outputs. If a program can produce results that are more similar to the reference outputs, it gets a higher score and can be considered more useful.


We originally attempted to use a code similarity approach, \ie comparing the key components in the generated program with those in oracle programs. However, this approach cannot handle the diverse ways of implementing a task, especially when considering other development frameworks. Therefore, we arrived at two metrics that are independent of the program structure, \textbf{Executable Rate} and \textbf{Result Score}.

\textbf{Executable Rate} measures how likely the programs generated by a generator are to be executable. It is a direct metric that can be obtained by simply running the program in the simulated environment.



\textbf{Result score} measures the similarity between the output data of the generated programs and the oracle programs. Directly comparing the results using traditional sequence similarity analysis techniques is not effective given the randomness and diversity of sensing programs. We introduce a customized fuzzy comparison algorithm to obtain the result score.

Specifically, we collect the output streams of generated programs and target programs into a list of data items.
We define the output data from oracle program and the target program as $\mathbf{A} = [\text{Item}_1, \text{Item}_2, \ldots, \text{Item}_m]$ and $\mathbf{B} = [\text{Item}_1', \text{Item}_2',$ $\ldots, \text{Item}_n']$ respectively. Each item field's value is treated as a string (\texttt{str}). The comparison between the two lists can be viewed as a customized LCS (Longest Common Subsequence) \cite{LCS} problem. 

We first define an item similarity function $F(\text{Item}_i, \text{Item}_j')$, which uses $\text{Item}_i$ as a reference to compare the similarity of common fields between $\text{Item}_i$ and $\text{Item}_j'$. Note that $F$ is a weighted sum based on the field similarity function $f$:
\begin{equation}
F(\text{Item}_i, \text{Item}_j') = \frac{\sum_{k \in \mathcal{K}} w_k \cdot f(v_{ik}, v_{jk}')}{\sum_{k \in \mathcal{K}} w_k}
\end{equation}
where the $\mathcal{K}$ is the set of common fields between $\text{Item}_i$ and $\text{Item}_j'$.The $v_{ik}$ and $v_{jk}'$ are the values of field $k$ in $\text{Item}_i$ and $\text{Item}_j'$ respectively. The $f(v_{ik}, v_{jk}')$ is the similarity function comparing two strings $v_{ik}$ and $v_{jk}'$, returning a real number in the range $[0, 1]$, common choices include BLEU \cite{papineni2002bleu} or Edit Distance (ED) \cite{navarro2001guided}. The $w_k$ is the weight for field $k$. One can set small weights for less important fields, and larger weights to highlight the essential fields.
We use dynamic programming to calculate the maximum similarity score \(\text{Similarity}(\mathbf{A}_i, \mathbf{B}_i)\) between the two item lists $\mathbf{A}$ and $\mathbf{B}$.

For multi-stream output tasks, where \(\mathcal{A}\) and \(\mathcal{B}\) contain the same number of streams and the stream names are known, the similarity between the two sets of streams is defined as the weighted average of the similarities between corresponding streams with the same name. Let \(\mathbf{A}_i \in \mathcal{A}\) and \(\mathbf{B}_i \in \mathcal{B}\) represent corresponding streams. The set similarity \(S(\mathcal{A}, \mathcal{B})\) is given by:

\begin{equation}
    S(\mathcal{A}, \mathcal{B}) = \frac{\sum_{i=1}^{|\mathcal{A}|} w_i \cdot \text{Similarity}(\mathbf{A}_i, \mathbf{B}_i)}{\sum_{i=1}^{|\mathcal{A}|} w_i}
\end{equation}
where \(w_i\) is the weight for the pair of streams \(\mathbf{A}_i\) and \(\mathbf{B}_i\), typically based on the length of the lists. The final score $S(\mathcal{A}, \mathcal{B})$ is used as the \textbf{Result Score} metric to measure the quality of the generated program.



\section{Evaluation}


\subsection{Experimental Setup}

\textbf{Tasks and Metrics.} 
To evaluate the system’s ability to solve context-aware tasks in an end-to-end manner, we employ 133 tasks from the proposed benchmark and two metrics: \textbf{Executable Rate} and \textbf{Result Score}, as described in Section~\ref{sec:benchmark}. 
All field weights \( w_k \) in Equation 1 are set to 1, meaning all fields are considered equally important. Moreover, the field similarity function \( f \) is chosen as BLEU to compare field similarity. The weight \( w_{i} \) in Equation 3 is equal to the length of the sequence, namely the result is weighted according to the length of the stream.



\textbf{Baselines.} To the best of our knowledge, there is no existing end-to-end system that can be directly compared with. Therefore, we attempt to construct the baselines by adapting existing techniques. 
The first baseline category is \textbf{code-free approaches} (Represented as \textbf{GPT4} and \textbf{GPT4o} in Table \ref{tab: result_with_baseline}), where the foundation model processes data without generating code. We use state-of-the-art OpenAI GPT-4 and GPT-4o models to establish these baselines. 
The second category, \textbf{code-based approaches}, involves generating a program to handle the sensing task. 
We included two baselines in this category. One is to generate native Python code, the other uses a popular LLM agent development framework LangChain \cite{Chase_LangChain_2022}. We include basic document of the frameworks in the program generation query like ours, although their complex APIs cannot be included in detail. The LLM we use with the frameworks is GPT-4o (same as our approach), which contains pre-trained knowledge of Python and LangChain. 
It is important to note that \textbf{directly generating the programs based on existing frameworks is nearly impossible to achieve the complex sensing goals}. Thus, we integrate the same stream-based abstraction of \sys to these baselines, allowing the code to access our stream data with \texttt{get()} and \texttt{put()} functions.





\subsection{Generation Quality}

\begin{table*}[ht]
    \centering
    \caption{The generation quality of different generators on the \bench benchmark.}
    \label{tab: result_with_baseline}
    \resizebox{\textwidth}{!}{
    \begin{tabular}{c|cccccc|cccccc}
    \toprule
    
    \multicolumn{1}{c|}{\textbf{Method}} & \multicolumn{6}{c|}{\textbf{Executable Rate }} & \multicolumn{6}{c}{\textbf{Result Score}}\\
    
    Repetitions  & \multicolumn{2}{c|}{\textbf{@1}} & \multicolumn{2}{c|}{\textbf{@3}} & \multicolumn{2}{c|}{\textbf{@5}}  & \multicolumn{2}{c|}{\textbf{@1}} & \multicolumn{2}{c|}{\textbf{@3}} & \multicolumn{2}{c}{\textbf{@5}}  \\

    Example Number & 0 & \multicolumn{1}{c|}{1} & 0 & \multicolumn{1}{c|}{1} & 0 & \multicolumn{1}{c|}{1} & 0 & \multicolumn{1}{c|}{1} & 0 & \multicolumn{1}{c|}{1} & 0 & \multicolumn{1}{c}{1} \\
    
    \midrule
    \textbf{GPT4}  & -  & - & - & -  & - & - & 0.354 & - & 0.421 & -  & 0.453 & - \\
    \textbf{GPT4o}  & -  & - & - & -  & - & - & 0.306  & - & 0.439 & -  & 0.528 & - \\
    \hline
    \textbf{Python w/ Abs.}  & 93.2\%  & 98.5\% & 100\%  & 100\%   & 100\%  & 100\%  & 0.227  & 0.351 & 0.395 & 0.464  & 0.458 & 0.500 \\
    \textbf{LangChain w/ Abs.}  & 77.4\%  & 97.0\% & 98.5\% & 100\%  & 100\% & 100\% & 0.188  & 0.375 & 0.320 & 0.492  & 0.377 & 0.526 \\
    \hline
    \textbf{Ours w/o feedback}  & 72.2\%  & 95.6\% & 93.2\% & 100\%  & 97.0\% & 100\% & 0.307 & 0.440 & 0.484 & 0.593  & 0.529 & 0.621 \\
    \textbf{Ours w/ feedback}  & 94.0\%  & 92.5\% & 100\% & 100\%  & 100\% & 100\% & \textbf{0.442} & \textbf{0.468} & \textbf{0.586} & \textbf{0.650} & \textbf{0.638} & \textbf{0.700} \\

    \bottomrule
    \end{tabular}
    }
\end{table*}

We first evaluate the context-aware program generation performance of \sys and various baselines. For each method, we assess two scenarios: without a reference example and with one reference example. Our method offers two approaches: the complete method (as shown in Chapter 4) and a one-time generation approach without the feedback process (Ours w/o feedback). We also record the best result each method could achieve for each task by generating multiple times (N times @N). The final results are shown in Table \ref{tab: result_with_baseline}. Since LLMs do not generate code, they do not have an executable rate metric and do not use a reference example. 

Generating Python code achieves a high executable rate regardless of the presence of examples, thanks to the extensive pre-trained knowledge in the GPT-4o generator. In contrast, using LangChain framework, based on the same model, has a lower success rate due to the framework's complexity and less pre-trained knowledge, though it can still achieve higher scores through examples and multiple attempts. 
\sys also faces challenges related to complexity and a lack of pre-trained knowledge. Ours w/o feedback method still encounters issues with execution failures despite multiple attempts, but the reference example can effectively address this problem. Ours w/ feedback method maintains a high success rate both with and without examples.

Code-free approaches can also achieve high result scores, demonstrating their powerful capabilities. However, their inability to generate code weakens their long-term stability in context-aware applications. Upon analysis, they also handle numerical tasks less precisely and perform worse on more complex tasks. Without examples, the Python method performs better than LangChain, again due to its pre-trained knowledge. However, with examples, LangChain surpasses Python, primarily because LangChain's main APIs are more highly abstracted. Our methods achieve the best results in various scenarios. Ours w/o feedback method provides a performance improvement with similar cost and latency compared to the baselines, while the Ours w/ feedback method significantly outperforms the baselines.

\label{sec:result_for_task_tag}

\begin{figure}
    \centering
    \includegraphics[width=0.9\linewidth]{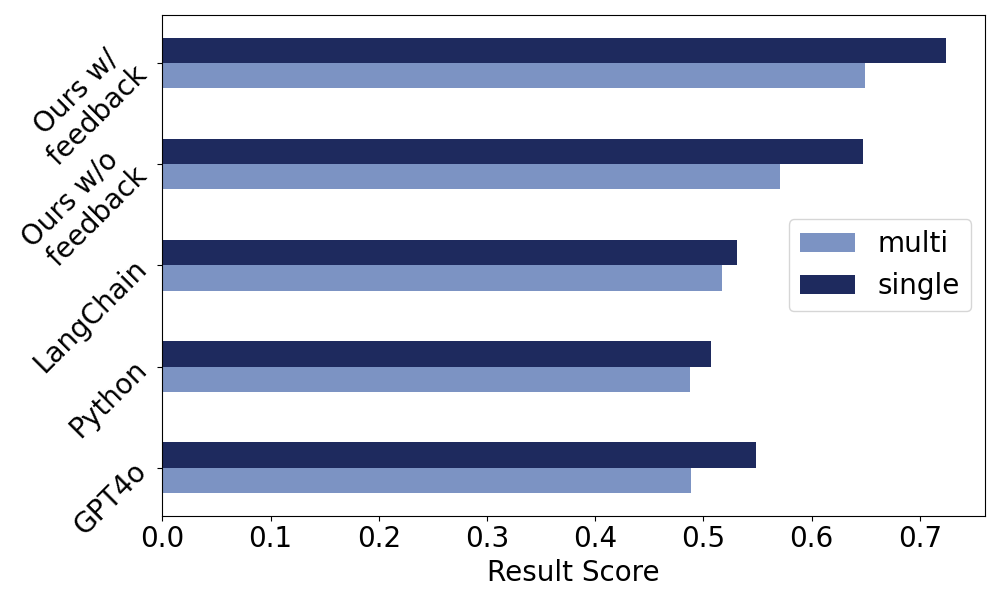}
    
    \caption{The generation quality of different approaches on multi- and single-function tasks with one reference example @5.}
    \label{fig:baseline_result_on_different_difficulty}
\end{figure}

\textbf{Performance on different levels of difficulty.}
Next, we analyze the performance of our method on tasks with varying levels of complexity. Our benchmark consists of 133 tasks in total, comprising 87 relatively simple single-step tasks and 46 more complex multi-step tasks. Figure \ref{fig:baseline_result_on_different_difficulty} shows the result scores of various methods on both single-step and multi-step tasks. It is evident that all methods achieve higher scores on single-step tasks compared to multi-step tasks. Additionally, it can be observed that GPT-4o performs similarly to Python and LangChain on complex tasks but significantly outperforms them on simple tasks. This discrepancy boosts GPT-4o’s average score, resulting in a total score in Table \ref{tab: result_with_baseline} that is higher than those of Python and LangChain. In contrast, our methods— both with and without feedback— significantly outperform other methods on both simple and complex tasks.

\subsection{ Latency and Cost Analysis }

\begin{table}
    \centering
    \caption{Average latency and token cost of different generators.}
   
    \label{tab:latency_and_token_cost}
    \resizebox{\columnwidth}{!}{
    \begin{tabular}{cccc}
        \toprule
        \textbf{Method} & \textbf{Latency (s)} & \textbf{Token (k)} & \textbf{Loop} \\
        \midrule
        Python & 5.61 & 0.70 & - \\
        LangChain & 6.19 & 0.77 & - \\
        Ours w/o feedback & 6.63 & 2.64 & - \\
        Ours  & 69.5 (19.3 per loop) & 19.6 (5.4 per loop) & 3.6 \\
        
        \bottomrule
    \end{tabular}
    }
\end{table}


We analyze the average time and token overhead required to generate code across all tasks, with the results shown in Table \ref{tab:latency_and_token_cost}. The Python, LangChain, and Ours w/o feedback methods are all single-generation methods, with their latency primarily dependent on the wait time for querying the cloud LLM. Since the prompt includes the framework manual, the token overhead is relatively higher. 

Ours w/ feedback, on average, requires 3.6 iterations to complete the generation task, leading to greater latency and token overhead. The number of iterations varies depending on the difficulty of the task. The main factors contributing to the latency are the waiting time for the cloud LLM and the time required for the sandbox to simulate the current code. The increase in token overhead mainly comes from the growing number of queries during iterations, which include the thought process, code, and feedback in each iteration. 

Overall, Ours w/ feedback trades higher overhead for the best results, while for simpler tasks, the non-iterative feedback method can still achieve performance that exceeds the baselines.



\subsection{ Fine-grained Performance Analysis }
We conduct ablation experiments to investigate the role of each component in the generation process. We primarily explore three questions: 1. What influence does feedback have? 2. What influence does the reference example have? 3. What influence do different types of information in feedback have?

\subsubsection{The influence of the feedback mechanism}
\begin{figure}
    \centering
    \includegraphics[width=0.9\linewidth]{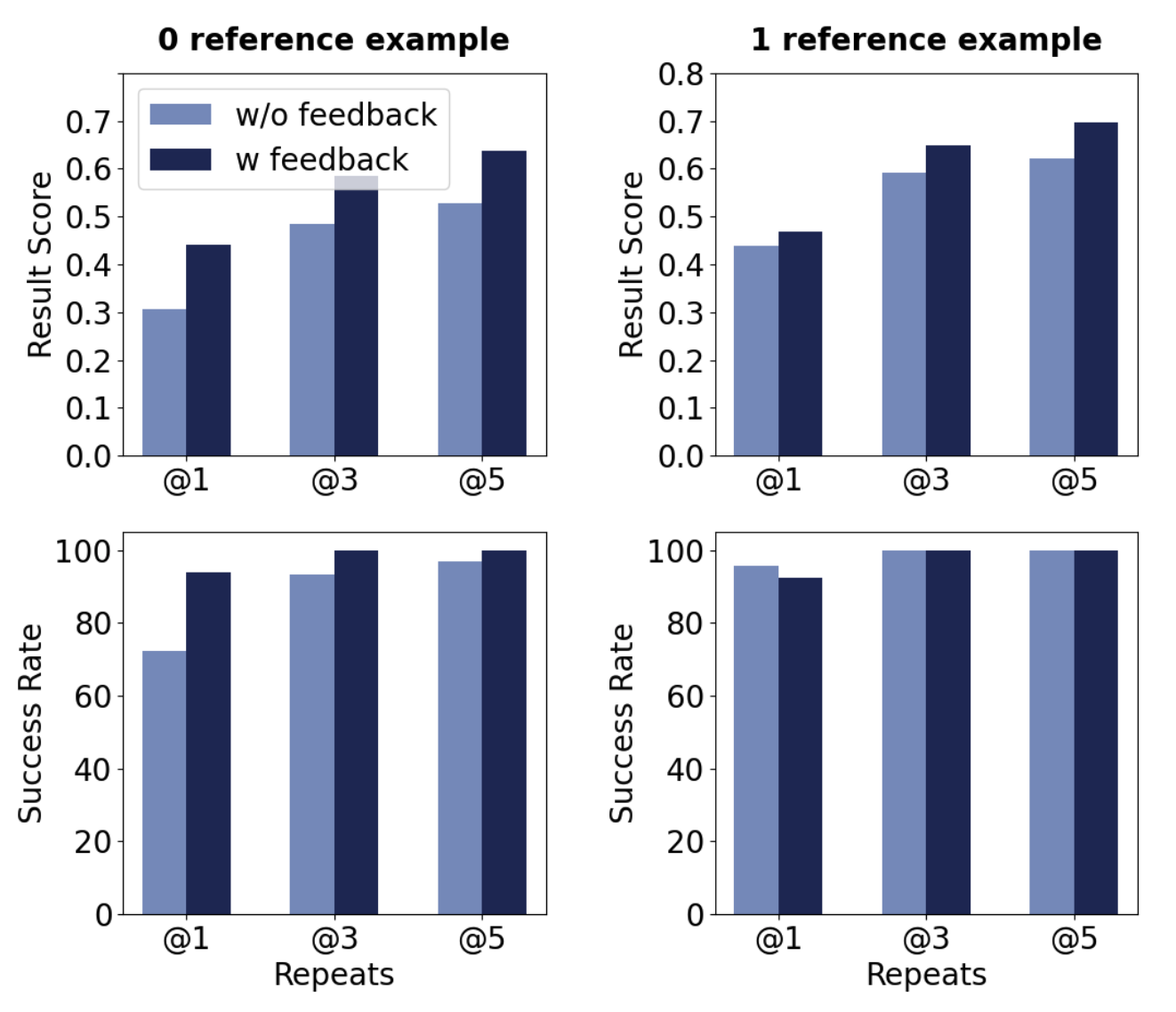}
    
    \caption{The influence of the feedback mechanism on the generation quality of \sys. }
    \label{fig:ablation_for_feedback}
\end{figure}

Figure \ref{fig:ablation_for_feedback} illustrates the influence of feedback on result scores and success rates in both scenarios with and without reference examples. It is evident that the dark-colored methods with feedback consistently outperform the methods without feedback in almost all aspects. Additionally, in the case of zero reference examples, feedback proves to be even more crucial, significantly improving the result score of the generated code and maintaining a higher success rate for the code.

\subsubsection{The influence of in-context examples.}
\begin{figure}
    \centering
    \includegraphics[width=0.9\linewidth]{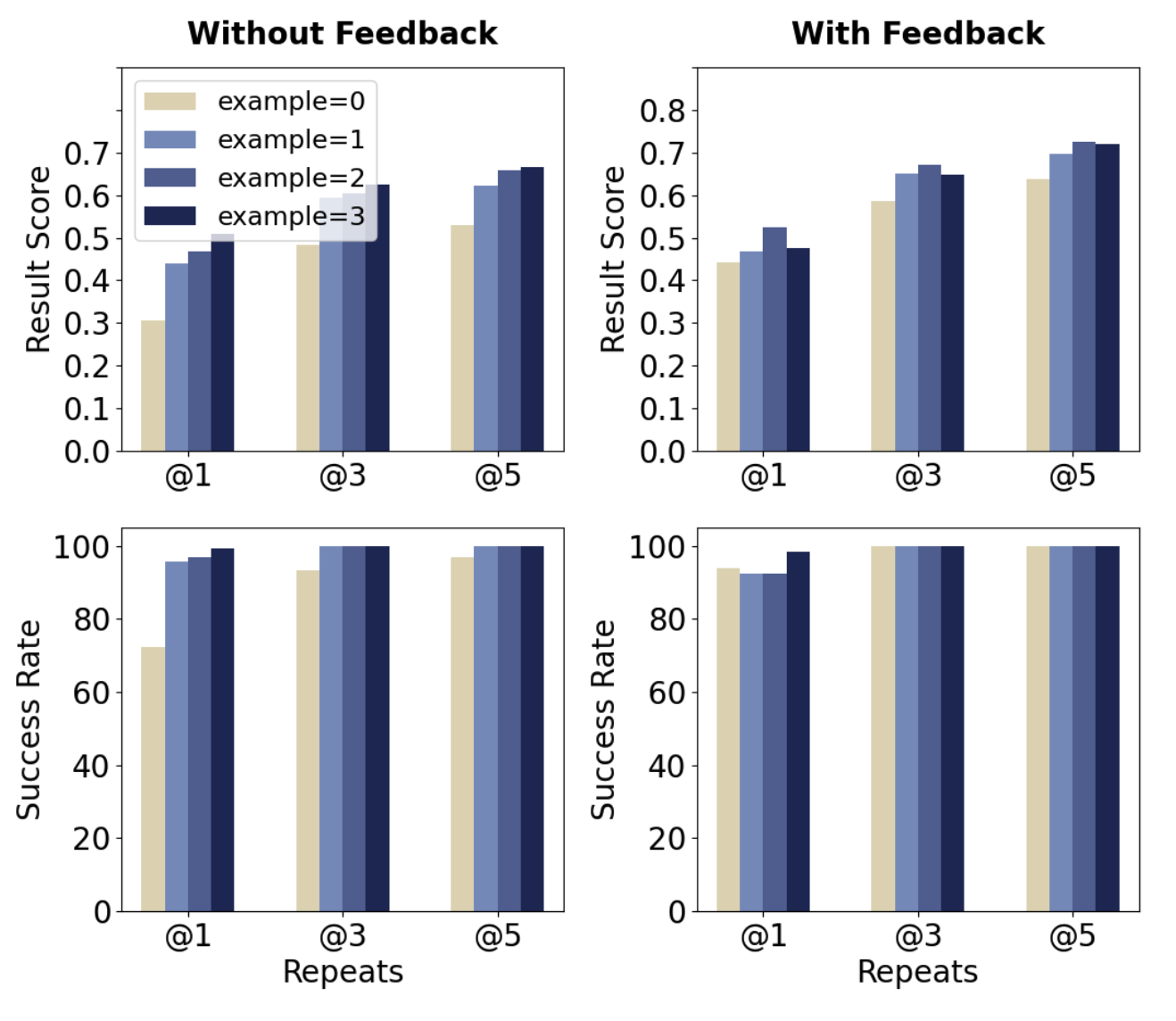}

    \caption{The influence of the in-context example number on the generation quality of \sys.}
    \label{fig:ablation_for_example_number}
\end{figure}

Figure \ref{fig:ablation_for_example_number} shows the effect of different numbers of examples on result scores and success rates in both feedback and no-feedback scenarios. As the color of the bars deepens, it represents an increase in the number of examples, and most of the scores rise accordingly. Similarly, the influence of examples is more significant in methods without feedback, leading to more noticeable improvements in both success rates and result scores. 
However, in methods with feedback, having too many examples can cause confusion and sometimes worsen the results. 

\subsubsection{The influence of different components in the feedback}

In the feedback process, besides the reference example, there are three main components: target stream output, error message, and logs. We conduct experiments by omitting each component separately, and the results are shown in Table \ref{tab:ablation_feedback_parts}. The absence of any part leads to a performance drop. 

In our analysis, we observe that each component plays a distinct role: the error message indicates hard errors in the code, the output provides the result of the code, and the logs allow the LLM to debug by printing key information. 

When the output is missing, it causes the most damage to the result because the output is the final outcome, which not only increases the likelihood of the program not executable but also decreases the result score of the executable code. The absence of error messages also influences the executable rate, but it has less effect on the result scores of executable code. When logs are omitted, there is a higher executable rate, but the result score decreases. This is because the generation process lacks the print debugging step, which simplifies the iteration process and makes the code easier to execute, but the overall code quality declines, leading to lower scores.


\begin{table}[ht]
    \centering
    \caption{The influence of different components in the feedback on the generation quality.} 
    
    \label{tab:ablation_feedback_parts}
    \resizebox{\columnwidth}{!}{
    \begin{tabular}{c|cc|cc}
    \toprule
    
    \multicolumn{1}{c|}{\textbf{Method}} & \multicolumn{2}{c|}{\textbf{Success Rate}} & \multicolumn{2}{c}{\textbf{Result Similarity Rate}}\\
    
    0 example  & @1  & $\Delta$  & @1 & $\Delta$  \\
    
    \midrule
    Ours w/o Output  & 70.7\%  & 23.3\% $\downarrow$ & 0.317 & 0.125 $\downarrow$ \\
    Ours w/o Error  & 71.4\%  & 22.6\% $\downarrow$ & 0.410 & 0.032 $\downarrow$ \\
    Ours w/o Logs  & 95.5\%  & 1.5\% $\uparrow$ & 0.427 & 0.015 $\downarrow$ \\
    \hline
    Ours  & \multicolumn{2}{c|}{ \textbf{94.0\%}}  & \multicolumn{2}{c}{ \textbf{0.442}} \\
    \bottomrule
    \end{tabular}
    }
    \end{table}



        







\section{Conclusion}

We have presented an end-to-end system named \sys for automatically generating context-sensing programs based on natural language. The framework features an easy-to-use stream-based programming interface and a feedback-guided query refinement method to make LLM-based generation easier and more precise. 



\bibliographystyle{ACM-Reference-Format}
\bibliography{references}


\begin{thebibliography}{53}


\ifx \showCODEN    \undefined \def \showCODEN     #1{\unskip}     \fi
\ifx \showDOI      \undefined \def \showDOI       #1{#1}\fi
\ifx \showISBNx    \undefined \def \showISBNx     #1{\unskip}     \fi
\ifx \showISBNxiii \undefined \def \showISBNxiii  #1{\unskip}     \fi
\ifx \showISSN     \undefined \def \showISSN      #1{\unskip}     \fi
\ifx \showLCCN     \undefined \def \showLCCN      #1{\unskip}     \fi
\ifx \shownote     \undefined \def \shownote      #1{#1}          \fi
\ifx \showarticletitle \undefined \def \showarticletitle #1{#1}   \fi
\ifx \showURL      \undefined \def \showURL       {\relax}        \fi
\providecommand\bibfield[2]{#2}
\providecommand\bibinfo[2]{#2}
\providecommand\natexlab[1]{#1}
\providecommand\showeprint[2][]{arXiv:#2}

\bibitem[Aditya et~al\mbox{.}(2016)]%
        {aditya2016pic}
\bibfield{author}{\bibinfo{person}{Paarijaat Aditya}, \bibinfo{person}{Rijurekha Sen}, \bibinfo{person}{Peter Druschel}, \bibinfo{person}{Seong Joon~Oh}, \bibinfo{person}{Rodrigo Benenson}, \bibinfo{person}{Mario Fritz}, \bibinfo{person}{Bernt Schiele}, \bibinfo{person}{Bobby Bhattacharjee}, {and} \bibinfo{person}{Tong~Tong Wu}.} \bibinfo{year}{2016}\natexlab{}.
\newblock \showarticletitle{I-pic: A platform for privacy-compliant image capture}. In \bibinfo{booktitle}{\emph{Proceedings of the 14th annual international conference on mobile systems, applications, and services}}. \bibinfo{pages}{235--248}.
\newblock


\bibitem[arXiv.org submitters(2024)]%
        {arxiv}
\bibfield{author}{\bibinfo{person}{arXiv.org submitters}.} \bibinfo{year}{2024}\natexlab{}.
\newblock \bibinfo{title}{arXiv Dataset}.
\newblock
\newblock
\urldef\tempurl%
\url{https://doi.org/10.34740/KAGGLE/DSV/7548853}
\showDOI{\tempurl}


\bibitem[Arzt et~al\mbox{.}(2014)]%
        {arzt2014flowdroid}
\bibfield{author}{\bibinfo{person}{Steven Arzt}, \bibinfo{person}{Siegfried Rasthofer}, \bibinfo{person}{Christian Fritz}, \bibinfo{person}{Eric Bodden}, \bibinfo{person}{Alexandre Bartel}, \bibinfo{person}{Jacques Klein}, \bibinfo{person}{Yves Le~Traon}, \bibinfo{person}{Damien Octeau}, {and} \bibinfo{person}{Patrick McDaniel}.} \bibinfo{year}{2014}\natexlab{}.
\newblock \showarticletitle{Flowdroid: Precise context, flow, field, object-sensitive and lifecycle-aware taint analysis for android apps}.
\newblock \bibinfo{journal}{\emph{ACM sigplan notices}} \bibinfo{volume}{49}, \bibinfo{number}{6} (\bibinfo{year}{2014}), \bibinfo{pages}{259--269}.
\newblock


\bibitem[Banken et~al\mbox{.}(2018)]%
        {banken2018debugging}
\bibfield{author}{\bibinfo{person}{Herman Banken}, \bibinfo{person}{Erik Meijer}, {and} \bibinfo{person}{Georgios Gousios}.} \bibinfo{year}{2018}\natexlab{}.
\newblock \showarticletitle{Debugging data flows in reactive programs}. In \bibinfo{booktitle}{\emph{Proceedings of the 40th international conference on software engineering}}. \bibinfo{pages}{752--763}.
\newblock


\bibitem[Chase(2022)]%
        {Chase_LangChain_2022}
\bibfield{author}{\bibinfo{person}{Harrison Chase}.} \bibinfo{year}{2022}\natexlab{}.
\newblock \bibinfo{booktitle}{\emph{{LangChain}}}.
\newblock
\urldef\tempurl%
\url{https://github.com/hwchase17/langchain}
\showURL{%
\tempurl}


\bibitem[Chen et~al\mbox{.}(2021)]%
        {HumanEval}
\bibfield{author}{\bibinfo{person}{Mark Chen}, \bibinfo{person}{Jerry Tworek}, \bibinfo{person}{Heewoo Jun}, \bibinfo{person}{Qiming Yuan}, \bibinfo{person}{Henrique~Ponde de Oliveira~Pinto}, \bibinfo{person}{Jared Kaplan}, \bibinfo{person}{Harri Edwards}, \bibinfo{person}{Yuri Burda}, \bibinfo{person}{Nicholas Joseph}, \bibinfo{person}{Greg Brockman}, \bibinfo{person}{Alex Ray}, \bibinfo{person}{Raul Puri}, \bibinfo{person}{Gretchen Krueger}, \bibinfo{person}{Michael Petrov}, \bibinfo{person}{Heidy Khlaaf}, \bibinfo{person}{Girish Sastry}, \bibinfo{person}{Pamela Mishkin}, \bibinfo{person}{Brooke Chan}, \bibinfo{person}{Scott Gray}, \bibinfo{person}{Nick Ryder}, \bibinfo{person}{Mikhail Pavlov}, \bibinfo{person}{Alethea Power}, \bibinfo{person}{Lukasz Kaiser}, \bibinfo{person}{Mohammad Bavarian}, \bibinfo{person}{Clemens Winter}, \bibinfo{person}{Philippe Tillet}, \bibinfo{person}{Felipe~Petroski Such}, \bibinfo{person}{Dave Cummings}, \bibinfo{person}{Matthias Plappert}, \bibinfo{person}{Fotios Chantzis},
  \bibinfo{person}{Elizabeth Barnes}, \bibinfo{person}{Ariel Herbert-Voss}, \bibinfo{person}{William~Hebgen Guss}, \bibinfo{person}{Alex Nichol}, \bibinfo{person}{Alex Paino}, \bibinfo{person}{Nikolas Tezak}, \bibinfo{person}{Jie Tang}, \bibinfo{person}{Igor Babuschkin}, \bibinfo{person}{Suchir Balaji}, \bibinfo{person}{Shantanu Jain}, \bibinfo{person}{William Saunders}, \bibinfo{person}{Christopher Hesse}, \bibinfo{person}{Andrew~N. Carr}, \bibinfo{person}{Jan Leike}, \bibinfo{person}{Josh Achiam}, \bibinfo{person}{Vedant Misra}, \bibinfo{person}{Evan Morikawa}, \bibinfo{person}{Alec Radford}, \bibinfo{person}{Matthew Knight}, \bibinfo{person}{Miles Brundage}, \bibinfo{person}{Mira Murati}, \bibinfo{person}{Katie Mayer}, \bibinfo{person}{Peter Welinder}, \bibinfo{person}{Bob McGrew}, \bibinfo{person}{Dario Amodei}, \bibinfo{person}{Sam McCandlish}, \bibinfo{person}{Ilya Sutskever}, {and} \bibinfo{person}{Wojciech Zaremba}.} \bibinfo{year}{2021}\natexlab{}.
\newblock \bibinfo{title}{Evaluating Large Language Models Trained on Code}.
\newblock
\newblock
\showeprint[arxiv]{2107.03374}~[cs.LG]
\urldef\tempurl%
\url{https://arxiv.org/abs/2107.03374}
\showURL{%
\tempurl}


\bibitem[Chen and Kan(2012)]%
        {SMS}
\bibfield{author}{\bibinfo{person}{Tao Chen} {and} \bibinfo{person}{Min-Yen Kan}.} \bibinfo{year}{2012}\natexlab{}.
\newblock \showarticletitle{Creating a live, public short message service corpus: the NUS SMS corpus}.
\newblock \bibinfo{journal}{\emph{Language Resources and Evaluation}} (\bibinfo{date}{aug} \bibinfo{year}{2012}).
\newblock
\showISSN{1574-0218}
\urldef\tempurl%
\url{https://doi.org/10.1007/s10579-012-9197-9}
\showDOI{\tempurl}


\bibitem[Cukierski(2016)]%
        {emails}
\bibfield{author}{\bibinfo{person}{Will Cukierski}.} \bibinfo{year}{2016}\natexlab{}.
\newblock \bibinfo{title}{The Enron Email Dataset}.
\newblock
\newblock
\urldef\tempurl%
\url{https://www.kaggle.com/datasets/wcukierski/enron-email-dataset?select=emails.csv}
\showURL{%
\tempurl}


\bibitem[Culler et~al\mbox{.}(1986)]%
        {culler1986dataflow}
\bibfield{author}{\bibinfo{person}{David~E Culler} {et~al\mbox{.}}} \bibinfo{year}{1986}\natexlab{}.
\newblock \showarticletitle{Dataflow architectures}.
\newblock  (\bibinfo{year}{1986}).
\newblock


\bibitem[Dai et~al\mbox{.}(2024)]%
        {dai2024advancing}
\bibfield{author}{\bibinfo{person}{Shenghong Dai}, \bibinfo{person}{Shiqi Jiang}, \bibinfo{person}{Yifan Yang}, \bibinfo{person}{Ting Cao}, \bibinfo{person}{Mo Li}, \bibinfo{person}{Suman Banerjee}, {and} \bibinfo{person}{Lili Qiu}.} \bibinfo{year}{2024}\natexlab{}.
\newblock \showarticletitle{Advancing Multi-Modal Sensing Through Expandable Modality Alignment}.
\newblock \bibinfo{journal}{\emph{arXiv preprint arXiv:2407.17777}} (\bibinfo{year}{2024}).
\newblock


\bibitem[de~la Sierra(2018)]%
        {wifi}
\bibfield{author}{\bibinfo{person}{Jesús~Martín de~la Sierra}.} \bibinfo{year}{2018}\natexlab{}.
\newblock \bibinfo{title}{WiFi data}.
\newblock
\newblock
\urldef\tempurl%
\url{https://www.kaggle.com/datasets/jmartindelasierra/wifi-data?select=wifi_data.csv}
\showURL{%
\tempurl}


\bibitem[De~Montjoye et~al\mbox{.}(2014)]%
        {de2014openpds}
\bibfield{author}{\bibinfo{person}{Yves-Alexandre De~Montjoye}, \bibinfo{person}{Erez Shmueli}, \bibinfo{person}{Samuel~S Wang}, {and} \bibinfo{person}{Alex~Sandy Pentland}.} \bibinfo{year}{2014}\natexlab{}.
\newblock \showarticletitle{openpds: Protecting the privacy of metadata through safeanswers}.
\newblock \bibinfo{journal}{\emph{PloS one}} \bibinfo{volume}{9}, \bibinfo{number}{7} (\bibinfo{year}{2014}), \bibinfo{pages}{e98790}.
\newblock


\bibitem[Deka et~al\mbox{.}(2017)]%
        {rico}
\bibfield{author}{\bibinfo{person}{Biplab Deka}, \bibinfo{person}{Zifeng Huang}, \bibinfo{person}{Chad Franzen}, \bibinfo{person}{Joshua Hibschman}, \bibinfo{person}{Daniel Afergan}, \bibinfo{person}{Yang Li}, \bibinfo{person}{Jeffrey Nichols}, {and} \bibinfo{person}{Ranjitha Kumar}.} \bibinfo{year}{2017}\natexlab{}.
\newblock \showarticletitle{Rico: A Mobile App Dataset for Building Data-Driven Design Applications}. In \bibinfo{booktitle}{\emph{Proceedings of the 30th Annual ACM Symposium on User Interface Software and Technology}} (Qu\'{e}bec City, QC, Canada) \emph{(\bibinfo{series}{UIST '17})}. \bibinfo{publisher}{Association for Computing Machinery}, \bibinfo{address}{New York, NY, USA}, \bibinfo{pages}{845–854}.
\newblock
\showISBNx{9781450349819}
\urldef\tempurl%
\url{https://doi.org/10.1145/3126594.3126651}
\showDOI{\tempurl}


\bibitem[Eight(2022)]%
        {twitter}
\bibfield{author}{\bibinfo{person}{Figure Eight}.} \bibinfo{year}{2022}\natexlab{}.
\newblock \bibinfo{title}{Twitter US Airline Sentiment}.
\newblock
\newblock
\urldef\tempurl%
\url{https://www.kaggle.com/datasets/crowdflower/twitter-airline-sentiment?select=Tweets.csv}
\showURL{%
\tempurl}


\bibitem[Enck et~al\mbox{.}(2014)]%
        {enck2014taintdroid}
\bibfield{author}{\bibinfo{person}{William Enck}, \bibinfo{person}{Peter Gilbert}, \bibinfo{person}{Seungyeop Han}, \bibinfo{person}{Vasant Tendulkar}, \bibinfo{person}{Byung-Gon Chun}, \bibinfo{person}{Landon~P Cox}, \bibinfo{person}{Jaeyeon Jung}, \bibinfo{person}{Patrick McDaniel}, {and} \bibinfo{person}{Anmol~N Sheth}.} \bibinfo{year}{2014}\natexlab{}.
\newblock \showarticletitle{Taintdroid: an information-flow tracking system for realtime privacy monitoring on smartphones}.
\newblock \bibinfo{journal}{\emph{ACM Transactions on Computer Systems (TOCS)}} \bibinfo{volume}{32}, \bibinfo{number}{2} (\bibinfo{year}{2014}), \bibinfo{pages}{1--29}.
\newblock


\bibitem[Ernst et~al\mbox{.}(2014)]%
        {ernst2014collaborative}
\bibfield{author}{\bibinfo{person}{Michael~D Ernst}, \bibinfo{person}{Ren{\'e} Just}, \bibinfo{person}{Suzanne Millstein}, \bibinfo{person}{Werner Dietl}, \bibinfo{person}{Stuart Pernsteiner}, \bibinfo{person}{Franziska Roesner}, \bibinfo{person}{Karl Koscher}, \bibinfo{person}{Paulo~Barros Barros}, \bibinfo{person}{Ravi Bhoraskar}, \bibinfo{person}{Seungyeop Han}, {et~al\mbox{.}}} \bibinfo{year}{2014}\natexlab{}.
\newblock \showarticletitle{Collaborative verification of information flow for a high-assurance app store}. In \bibinfo{booktitle}{\emph{Proceedings of the 2014 ACM SIGSAC Conference on Computer and Communications Security}}. \bibinfo{pages}{1092--1104}.
\newblock


\bibitem[Ferreira et~al\mbox{.}(2015)]%
        {ferreira2015aware}
\bibfield{author}{\bibinfo{person}{Denzil Ferreira}, \bibinfo{person}{Vassilis Kostakos}, {and} \bibinfo{person}{Anind~K Dey}.} \bibinfo{year}{2015}\natexlab{}.
\newblock \showarticletitle{AWARE: mobile context instrumentation framework}.
\newblock \bibinfo{journal}{\emph{Frontiers in ICT}}  \bibinfo{volume}{2} (\bibinfo{year}{2015}), \bibinfo{pages}{6}.
\newblock


\bibitem[Gawlik(2016)]%
        {stock}
\bibfield{author}{\bibinfo{person}{Dominik Gawlik}.} \bibinfo{year}{2016}\natexlab{}.
\newblock \bibinfo{title}{New York Stock Exchange}.
\newblock
\newblock
\urldef\tempurl%
\url{https://www.kaggle.com/datasets/dgawlik/nyse?select=prices-split-adjusted.csv}
\showURL{%
\tempurl}


\bibitem[Grauman et~al\mbox{.}(2022)]%
        {ego4d}
\bibfield{author}{\bibinfo{person}{Kristen Grauman}, \bibinfo{person}{Andrew Westbury}, \bibinfo{person}{Eugene Byrne}, \bibinfo{person}{Zachary Chavis}, \bibinfo{person}{Antonino Furnari}, \bibinfo{person}{Rohit Girdhar}, \bibinfo{person}{Jackson Hamburger}, \bibinfo{person}{Hao Jiang}, \bibinfo{person}{Miao Liu}, \bibinfo{person}{Xingyu Liu}, \bibinfo{person}{Miguel Martin}, \bibinfo{person}{Tushar Nagarajan}, \bibinfo{person}{Ilija Radosavovic}, \bibinfo{person}{Santhosh~Kumar Ramakrishnan}, \bibinfo{person}{Fiona Ryan}, \bibinfo{person}{Jayant Sharma}, \bibinfo{person}{Michael Wray}, \bibinfo{person}{Mengmeng Xu}, \bibinfo{person}{Eric~Zhongcong Xu}, \bibinfo{person}{Chen Zhao}, \bibinfo{person}{Siddhant Bansal}, \bibinfo{person}{Dhruv Batra}, \bibinfo{person}{Vincent Cartillier}, \bibinfo{person}{Sean Crane}, \bibinfo{person}{Tien Do}, \bibinfo{person}{Morrie Doulaty}, \bibinfo{person}{Akshay Erapalli}, \bibinfo{person}{Christoph Feichtenhofer}, \bibinfo{person}{Adriano Fragomeni},
  \bibinfo{person}{Qichen Fu}, \bibinfo{person}{Christian Fuegen}, \bibinfo{person}{Abrham Gebreselasie}, \bibinfo{person}{Cristina Gonzalez}, \bibinfo{person}{James Hillis}, \bibinfo{person}{Xuhua Huang}, \bibinfo{person}{Yifei Huang}, \bibinfo{person}{Wenqi Jia}, \bibinfo{person}{Weslie Khoo}, \bibinfo{person}{Jachym Kolar}, \bibinfo{person}{Satwik Kottur}, \bibinfo{person}{Anurag Kumar}, \bibinfo{person}{Federico Landini}, \bibinfo{person}{Chao Li}, \bibinfo{person}{Yanghao Li}, \bibinfo{person}{Zhenqiang Li}, \bibinfo{person}{Karttikeya Mangalam}, \bibinfo{person}{Raghava Modhugu}, \bibinfo{person}{Jonathan Munro}, \bibinfo{person}{Tullie Murrell}, \bibinfo{person}{Takumi Nishiyasu}, \bibinfo{person}{Will Price}, \bibinfo{person}{Paola~Ruiz Puentes}, \bibinfo{person}{Merey Ramazanova}, \bibinfo{person}{Leda Sari}, \bibinfo{person}{Kiran Somasundaram}, \bibinfo{person}{Audrey Southerland}, \bibinfo{person}{Yusuke Sugano}, \bibinfo{person}{Ruijie Tao}, \bibinfo{person}{Minh Vo}, \bibinfo{person}{Yuchen
  Wang}, \bibinfo{person}{Xindi Wu}, \bibinfo{person}{Takuma Yagi}, \bibinfo{person}{Yunyi Zhu}, \bibinfo{person}{Pablo Arbelaez}, \bibinfo{person}{David Crandall}, \bibinfo{person}{Dima Damen}, \bibinfo{person}{Giovanni~Maria Farinella}, \bibinfo{person}{Bernard Ghanem}, \bibinfo{person}{Vamsi~Krishna Ithapu}, \bibinfo{person}{C.~V. Jawahar}, \bibinfo{person}{Hanbyul Joo}, \bibinfo{person}{Kris Kitani}, \bibinfo{person}{Haizhou Li}, \bibinfo{person}{Richard Newcombe}, \bibinfo{person}{Aude Oliva}, \bibinfo{person}{Hyun~Soo Park}, \bibinfo{person}{James~M. Rehg}, \bibinfo{person}{Yoichi Sato}, \bibinfo{person}{Jianbo Shi}, \bibinfo{person}{Mike~Zheng Shou}, \bibinfo{person}{Antonio Torralba}, \bibinfo{person}{Lorenzo Torresani}, \bibinfo{person}{Mingfei Yan}, {and} \bibinfo{person}{Jitendra Malik}.} \bibinfo{year}{2022}\natexlab{}.
\newblock \showarticletitle{Ego4D: Around the {W}orld in 3,000 {H}ours of {E}gocentric {V}ideo}. In \bibinfo{booktitle}{\emph{IEEE/CVF Computer Vision and Pattern Recognition (CVPR)}}.
\newblock


\bibitem[Grecnik(2018)]%
        {world_gpsworld_gps}
\bibfield{author}{\bibinfo{person}{Grecnik}.} \bibinfo{year}{2018}\natexlab{}.
\newblock \bibinfo{title}{World capitals gps}.
\newblock
\newblock
\urldef\tempurl%
\url{https://www.kaggle.com/datasets/nikitagrec/world-capitals-gps?select=concap.csv}
\showURL{%
\tempurl}


\bibitem[Harari and Gosling(2023)]%
        {incontext_mobile_sensing_nature}
\bibfield{author}{\bibinfo{person}{Gabriella~M. Harari} {and} \bibinfo{person}{Samuel~D. Gosling}.} \bibinfo{year}{2023}\natexlab{}.
\newblock \showarticletitle{Understanding behaviours in context using mobile sensing}.
\newblock \bibinfo{journal}{\emph{Nature Reviews Psychology}} \bibinfo{volume}{2}, \bibinfo{number}{12} (\bibinfo{year}{2023}), \bibinfo{pages}{767--779}.
\newblock
\showISSN{2731-0574}
\urldef\tempurl%
\url{https://doi.org/10.1038/s44159-023-00235-3}
\showDOI{\tempurl}


\bibitem[Hirschberg(1977)]%
        {LCS}
\bibfield{author}{\bibinfo{person}{Daniel~S. Hirschberg}.} \bibinfo{year}{1977}\natexlab{}.
\newblock \showarticletitle{Algorithms for the Longest Common Subsequence Problem}.
\newblock \bibinfo{journal}{\emph{J. ACM}} \bibinfo{volume}{24}, \bibinfo{number}{4} (\bibinfo{date}{oct} \bibinfo{year}{1977}), \bibinfo{pages}{664–675}.
\newblock
\showISSN{0004-5411}
\urldef\tempurl%
\url{https://doi.org/10.1145/322033.322044}
\showDOI{\tempurl}


\bibitem[Hong et~al\mbox{.}(2024)]%
        {hong2024metagpt}
\bibfield{author}{\bibinfo{person}{Sirui Hong}, \bibinfo{person}{Mingchen Zhuge}, \bibinfo{person}{Jonathan Chen}, \bibinfo{person}{Xiawu Zheng}, \bibinfo{person}{Yuheng Cheng}, \bibinfo{person}{Jinlin Wang}, \bibinfo{person}{Ceyao Zhang}, \bibinfo{person}{Zili Wang}, \bibinfo{person}{Steven Ka~Shing Yau}, \bibinfo{person}{Zijuan Lin}, \bibinfo{person}{Liyang Zhou}, \bibinfo{person}{Chenyu Ran}, \bibinfo{person}{Lingfeng Xiao}, \bibinfo{person}{Chenglin Wu}, {and} \bibinfo{person}{J{\"u}rgen Schmidhuber}.} \bibinfo{year}{2024}\natexlab{}.
\newblock \showarticletitle{Meta{GPT}: Meta Programming for A Multi-Agent Collaborative Framework}. In \bibinfo{booktitle}{\emph{The Twelfth International Conference on Learning Representations}}.
\newblock
\urldef\tempurl%
\url{https://openreview.net/forum?id=VtmBAGCN7o}
\showURL{%
\tempurl}


\bibitem[Hossain(2024)]%
        {activities}
\bibfield{author}{\bibinfo{person}{Yamin Hossain}.} \bibinfo{year}{2024}\natexlab{}.
\newblock \bibinfo{title}{Fitness Track Daily Activity Dataset}.
\newblock
\newblock
\urldef\tempurl%
\url{https://www.kaggle.com/datasets/yaminh/fitness-track-daily-activity-dataset?select=Activity.csv}
\showURL{%
\tempurl}


\bibitem[Jin et~al\mbox{.}(2022)]%
        {jin2022peekaboo}
\bibfield{author}{\bibinfo{person}{Haojian Jin}, \bibinfo{person}{Gram Liu}, \bibinfo{person}{David Hwang}, \bibinfo{person}{Swarun Kumar}, \bibinfo{person}{Yuvraj Agarwal}, {and} \bibinfo{person}{Jason~I Hong}.} \bibinfo{year}{2022}\natexlab{}.
\newblock \showarticletitle{Peekaboo: A hub-based approach to enable transparency in data processing within smart homes}. In \bibinfo{booktitle}{\emph{2022 IEEE symposium on security and privacy (SP)}}. IEEE, \bibinfo{pages}{303--320}.
\newblock


\bibitem[kenneth ze(2023)]%
        {sleep_dataset}
\bibfield{author}{\bibinfo{person}{kenneth ze}.} \bibinfo{year}{2023}\natexlab{}.
\newblock \bibinfo{title}{Sleep health and lifestyle dataset}.
\newblock
\newblock
\urldef\tempurl%
\url{https://www.kaggle.com/datasets/informateur234/sleep-health-and-lifestyle-dataset?select=Sleep_health_and_lifestyle_dataset.csv}
\showURL{%
\tempurl}


\bibitem[Li et~al\mbox{.}(2017a)]%
        {li_imwut17_privacystreams}
\bibfield{author}{\bibinfo{person}{Yuanchun Li}, \bibinfo{person}{Fanglin Chen}, \bibinfo{person}{Toby Jia-Jun Li}, \bibinfo{person}{Yao Guo}, \bibinfo{person}{Gang Huang}, \bibinfo{person}{Matthew Fredrikson}, \bibinfo{person}{Yuvraj Agarwal}, {and} \bibinfo{person}{Jason~I Hong}.} \bibinfo{year}{2017}\natexlab{a}.
\newblock \showarticletitle{PrivacyStreams: Enabling transparency in personal data processing for mobile apps}.
\newblock \bibinfo{journal}{\emph{Proceedings of the ACM on Interactive, Mobile, Wearable and Ubiquitous Technologies}} \bibinfo{volume}{1}, \bibinfo{number}{3} (\bibinfo{year}{2017}), \bibinfo{pages}{76}.
\newblock


\bibitem[Li et~al\mbox{.}(2016)]%
        {li2016peruim}
\bibfield{author}{\bibinfo{person}{Yuanchun Li}, \bibinfo{person}{Yao Guo}, {and} \bibinfo{person}{Xiangqun Chen}.} \bibinfo{year}{2016}\natexlab{}.
\newblock \showarticletitle{Peruim: Understanding mobile application privacy with permission-ui mapping}. In \bibinfo{booktitle}{\emph{Proceedings of the 2016 ACM International Joint Conference on Pervasive and Ubiquitous Computing}}. \bibinfo{pages}{682--693}.
\newblock


\bibitem[Li et~al\mbox{.}(2017b)]%
        {dialog}
\bibfield{author}{\bibinfo{person}{Yanran Li}, \bibinfo{person}{Hui Su}, \bibinfo{person}{Xiaoyu Shen}, \bibinfo{person}{Wenjie Li}, \bibinfo{person}{Ziqiang Cao}, {and} \bibinfo{person}{Shuzi Niu}.} \bibinfo{year}{2017}\natexlab{b}.
\newblock \showarticletitle{DailyDialog: A Manually Labelled Multi-turn Dialogue Dataset}. In \bibinfo{booktitle}{\emph{Proceedings of The 8th International Joint Conference on Natural Language Processing (IJCNLP 2017)}}.
\newblock


\bibitem[Li et~al\mbox{.}(2024)]%
        {li2024personal_llm_agents}
\bibfield{author}{\bibinfo{person}{Yuanchun Li}, \bibinfo{person}{Hao Wen}, \bibinfo{person}{Weijun Wang}, \bibinfo{person}{Xiangyu Li}, \bibinfo{person}{Yizhen Yuan}, \bibinfo{person}{Guohong Liu}, \bibinfo{person}{Jiacheng Liu}, \bibinfo{person}{Wenxing Xu}, \bibinfo{person}{Xiang Wang}, \bibinfo{person}{Yi Sun}, \bibinfo{person}{Rui Kong}, \bibinfo{person}{Yile Wang}, \bibinfo{person}{Hanfei Geng}, \bibinfo{person}{Jian Luan}, \bibinfo{person}{Xuefeng Jin}, \bibinfo{person}{Zilong Ye}, \bibinfo{person}{Guanjing Xiong}, \bibinfo{person}{Fan Zhang}, \bibinfo{person}{Xiang Li}, \bibinfo{person}{Mengwei Xu}, \bibinfo{person}{Zhijun Li}, \bibinfo{person}{Peng Li}, \bibinfo{person}{Yang Liu}, \bibinfo{person}{Ya-Qin Zhang}, {and} \bibinfo{person}{Yunxin Liu}.} \bibinfo{year}{2024}\natexlab{}.
\newblock \showarticletitle{Personal LLM Agents: Insights and Survey about the Capability, Efficiency and Security}.
\newblock \bibinfo{journal}{\emph{arXiv preprint arXiv:2401.05459}} (\bibinfo{year}{2024}).
\newblock


\bibitem[Liu et~al\mbox{.}(2024)]%
        {liu2024chainstream}
\bibfield{author}{\bibinfo{person}{Jiacheng Liu}, \bibinfo{person}{Wenxing Xu}, {and} \bibinfo{person}{Yuanchun Li}.} \bibinfo{year}{2024}\natexlab{}.
\newblock \showarticletitle{ChainStream: A Stream-based LLM Agent Framework for Continuous Context Sensing and Sharing}. In \bibinfo{booktitle}{\emph{Proceedings of the Workshop on Edge and Mobile Foundation Models}}. \bibinfo{pages}{18--23}.
\newblock


\bibitem[Liu et~al\mbox{.}(2023)]%
        {llmhealthlearner}
\bibfield{author}{\bibinfo{person}{Xin Liu}, \bibinfo{person}{Daniel McDuff}, \bibinfo{person}{Geza Kovacs}, \bibinfo{person}{Isaac Galatzer-Levy}, \bibinfo{person}{Jacob Sunshine}, \bibinfo{person}{Jiening Zhan}, \bibinfo{person}{Ming-Zher Poh}, \bibinfo{person}{Shun Liao}, \bibinfo{person}{Paolo Di~Achille}, {and} \bibinfo{person}{Shwetak Patel}.} \bibinfo{year}{2023}\natexlab{}.
\newblock \showarticletitle{Large language models are few-shot health learners}.
\newblock \bibinfo{journal}{\emph{arXiv preprint arXiv:2305.15525}} (\bibinfo{year}{2023}).
\newblock


\bibitem[Malviya et~al\mbox{.}(2022)]%
        {malviya2022right}
\bibfield{author}{\bibinfo{person}{Vikas~Kumar Malviya}, \bibinfo{person}{Chee~Wei Leow}, \bibinfo{person}{Ashok Kasthuri}, \bibinfo{person}{Yan~Naing Tun}, \bibinfo{person}{Lwin~Khin Shar}, {and} \bibinfo{person}{Lingxiao Jiang}.} \bibinfo{year}{2022}\natexlab{}.
\newblock \showarticletitle{Right to Know, Right to Refuse: Towards UI Perception-Based Automated Fine-Grained Permission Controls for Android Apps}. In \bibinfo{booktitle}{\emph{Proceedings of the 37th IEEE/ACM International Conference on Automated Software Engineering}}. \bibinfo{pages}{1--6}.
\newblock


\bibitem[Melde(2020)]%
        {sphar}
\bibfield{author}{\bibinfo{person}{Alexander Melde}.} \bibinfo{year}{2020}\natexlab{}.
\newblock \showarticletitle{SPHAR: Surveillance Perspective Human Action Recognition Dataset}.
\newblock \bibinfo{journal}{\emph{GitHub repository}} (\bibinfo{year}{2020}).
\newblock
\urldef\tempurl%
\url{https://github.com/AlexanderMelde/SPHAR-Dataset}
\showURL{%
\tempurl}


\bibitem[Misra(2022)]%
        {news}
\bibfield{author}{\bibinfo{person}{Rishabh Misra}.} \bibinfo{year}{2022}\natexlab{}.
\newblock \bibinfo{title}{News Category Dataset}.
\newblock
\newblock
\urldef\tempurl%
\url{https://www.kaggle.com/datasets/rmisra/news-category-dataset?select=News_Category_Dataset_v3.json}
\showURL{%
\tempurl}


\bibitem[Navarro(2001)]%
        {navarro2001guided}
\bibfield{author}{\bibinfo{person}{Gonzalo Navarro}.} \bibinfo{year}{2001}\natexlab{}.
\newblock \showarticletitle{A guided tour to approximate string matching}.
\newblock \bibinfo{journal}{\emph{ACM computing surveys (CSUR)}} \bibinfo{volume}{33}, \bibinfo{number}{1} (\bibinfo{year}{2001}), \bibinfo{pages}{31--88}.
\newblock


\bibitem[of~Seattle(2021)]%
        {sea_building_energy}
\bibfield{author}{\bibinfo{person}{City of Seattle}.} \bibinfo{year}{2021}\natexlab{}.
\newblock \bibinfo{title}{SEA Building Energy Benchmarking}.
\newblock
\newblock
\urldef\tempurl%
\url{https://www.kaggle.com/datasets/city-of-seattle/sea-building-energy-benchmarking}
\showURL{%
\tempurl}


\bibitem[Okita et~al\mbox{.}(2023)]%
        {towardsllmforsensordata}
\bibfield{author}{\bibinfo{person}{Tsuyoshi Okita}, \bibinfo{person}{Kosuke Ukita}, \bibinfo{person}{Koki Matsuishi}, \bibinfo{person}{Masaharu Kagiyama}, \bibinfo{person}{Kodai Hirata}, {and} \bibinfo{person}{Asahi Miyazaki}.} \bibinfo{year}{2023}\natexlab{}.
\newblock \showarticletitle{Towards LLMs for Sensor Data: Multi-Task Self-Supervised Learning}. In \bibinfo{booktitle}{\emph{Adjunct Proceedings of the 2023 ACM International Joint Conference on Pervasive and Ubiquitous Computing \& the 2023 ACM International Symposium on Wearable Computing}} (Cancun, Quintana Roo, Mexico) \emph{(\bibinfo{series}{UbiComp/ISWC '23 Adjunct})}. \bibinfo{publisher}{Association for Computing Machinery}, \bibinfo{address}{New York, NY, USA}, \bibinfo{pages}{499–504}.
\newblock
\showISBNx{9798400702006}
\urldef\tempurl%
\url{https://doi.org/10.1145/3594739.3610745}
\showDOI{\tempurl}


\bibitem[Ouyang and Srivastava(2024)]%
        {ouyang2024llmsense}
\bibfield{author}{\bibinfo{person}{Xiaomin Ouyang} {and} \bibinfo{person}{Mani Srivastava}.} \bibinfo{year}{2024}\natexlab{}.
\newblock \showarticletitle{LLMSense: Harnessing LLMs for High-level Reasoning Over Spatiotemporal Sensor Traces}.
\newblock \bibinfo{journal}{\emph{arXiv preprint arXiv:2403.19857}} (\bibinfo{year}{2024}).
\newblock


\bibitem[Papineni et~al\mbox{.}(2002)]%
        {papineni2002bleu}
\bibfield{author}{\bibinfo{person}{Kishore Papineni}, \bibinfo{person}{Salim Roukos}, \bibinfo{person}{Todd Ward}, {and} \bibinfo{person}{Wei-Jing Zhu}.} \bibinfo{year}{2002}\natexlab{}.
\newblock \showarticletitle{Bleu: a method for automatic evaluation of machine translation}. In \bibinfo{booktitle}{\emph{Proceedings of the 40th annual meeting of the Association for Computational Linguistics}}. \bibinfo{pages}{311--318}.
\newblock


\bibitem[Patel et~al\mbox{.}(2024)]%
        {patel-etal-2024-evaluating}
\bibfield{author}{\bibinfo{person}{Arkil Patel}, \bibinfo{person}{Siva Reddy}, \bibinfo{person}{Dzmitry Bahdanau}, {and} \bibinfo{person}{Pradeep Dasigi}.} \bibinfo{year}{2024}\natexlab{}.
\newblock \showarticletitle{Evaluating In-Context Learning of Libraries for Code Generation}. In \bibinfo{booktitle}{\emph{Proceedings of the 2024 Conference of the North American Chapter of the Association for Computational Linguistics: Human Language Technologies (Volume 1: Long Papers)}}, \bibfield{editor}{\bibinfo{person}{Kevin Duh}, \bibinfo{person}{Helena Gomez}, {and} \bibinfo{person}{Steven Bethard}} (Eds.). \bibinfo{publisher}{Association for Computational Linguistics}, \bibinfo{address}{Mexico City, Mexico}, \bibinfo{pages}{2908--2926}.
\newblock
\urldef\tempurl%
\url{https://doi.org/10.18653/v1/2024.naacl-long.161}
\showDOI{\tempurl}


\bibitem[Patil(2024)]%
        {Weather_Data}
\bibfield{author}{\bibinfo{person}{Prasad Patil}.} \bibinfo{year}{2024}\natexlab{}.
\newblock \bibinfo{title}{Weather Data}.
\newblock
\newblock
\urldef\tempurl%
\url{https://www.kaggle.com/datasets/prasad22/weather-data?select=weather_data.csv}
\showURL{%
\tempurl}


\bibitem[Raj(2022)]%
        {GitHub_Dataset}
\bibfield{author}{\bibinfo{person}{Nikhil Raj}.} \bibinfo{year}{2022}\natexlab{}.
\newblock \bibinfo{title}{GitHub Dataset}.
\newblock
\newblock
\urldef\tempurl%
\url{https://www.kaggle.com/datasets/nikhil25803/github-dataset?select=github_dataset.csv}
\showURL{%
\tempurl}
\newblock
\shownote{Version 1 and Version 2}.


\bibitem[Raval et~al\mbox{.}(2016)]%
        {raval2016whatyoumark}
\bibfield{author}{\bibinfo{person}{Nisarg Raval}, \bibinfo{person}{Animesh Srivastava}, \bibinfo{person}{Ali Razeen}, \bibinfo{person}{Kiron Lebeck}, \bibinfo{person}{Ashwin Machanavajjhala}, {and} \bibinfo{person}{Lanodn~P Cox}.} \bibinfo{year}{2016}\natexlab{}.
\newblock \showarticletitle{What you mark is what apps see}. In \bibinfo{booktitle}{\emph{Proceedings of the 14th Annual International Conference on Mobile Systems, Applications, and Services}}. \bibinfo{pages}{249--261}.
\newblock


\bibitem[Sharma(2022)]%
        {maternal_helth}
\bibfield{author}{\bibinfo{person}{Dr.MB Sharma}.} \bibinfo{year}{2022}\natexlab{}.
\newblock \bibinfo{title}{Maternal Health Risk Data set}.
\newblock
\newblock
\urldef\tempurl%
\url{https://www.kaggle.com/datasets/drmbsharma/maternal-health-risk-data-set?select=Maternal_Risk.csv}
\showURL{%
\tempurl}


\bibitem[waltteri(2020)]%
        {desktop_ui}
\bibfield{author}{\bibinfo{person}{waltteri}.} \bibinfo{year}{2020}\natexlab{}.
\newblock \bibinfo{title}{desktop-ui-dataset}.
\newblock
\newblock
\urldef\tempurl%
\url{https://github.com/waltteri/desktop-ui-dataset}
\showURL{%
\tempurl}


\bibitem[Wei et~al\mbox{.}(2022)]%
        {wei2022cot}
\bibfield{author}{\bibinfo{person}{Jason Wei}, \bibinfo{person}{Xuezhi Wang}, \bibinfo{person}{Dale Schuurmans}, \bibinfo{person}{Maarten Bosma}, \bibinfo{person}{Fei Xia}, \bibinfo{person}{Ed Chi}, \bibinfo{person}{Quoc~V Le}, \bibinfo{person}{Denny Zhou}, {et~al\mbox{.}}} \bibinfo{year}{2022}\natexlab{}.
\newblock \showarticletitle{Chain-of-thought prompting elicits reasoning in large language models}.
\newblock \bibinfo{journal}{\emph{Advances in neural information processing systems}}  \bibinfo{volume}{35} (\bibinfo{year}{2022}), \bibinfo{pages}{24824--24837}.
\newblock


\bibitem[Xu et~al\mbox{.}(2024)]%
        {xu-etal-2024-penetrative}
\bibfield{author}{\bibinfo{person}{Huatao Xu}, \bibinfo{person}{Liying Han}, \bibinfo{person}{Qirui Yang}, \bibinfo{person}{Mo Li}, {and} \bibinfo{person}{Mani Srivastava}.} \bibinfo{year}{2024}\natexlab{}.
\newblock \showarticletitle{Penetrative {AI}: Making {LLM}s Comprehend the Physical World}. In \bibinfo{booktitle}{\emph{Findings of the Association for Computational Linguistics ACL 2024}}, \bibfield{editor}{\bibinfo{person}{Lun-Wei Ku}, \bibinfo{person}{Andre Martins}, {and} \bibinfo{person}{Vivek Srikumar}} (Eds.). \bibinfo{publisher}{Association for Computational Linguistics}, \bibinfo{address}{Bangkok, Thailand and virtual meeting}, \bibinfo{pages}{7324--7341}.
\newblock
\urldef\tempurl%
\url{https://aclanthology.org/2024.findings-acl.437}
\showURL{%
\tempurl}


\bibitem[Yang et~al\mbox{.}(2021)]%
        {taintstream2021}
\bibfield{author}{\bibinfo{person}{Chengxu Yang}, \bibinfo{person}{Yuanchun Li}, \bibinfo{person}{Mengwei Xu}, \bibinfo{person}{Zhenpeng Chen}, \bibinfo{person}{Yunxin Liu}, \bibinfo{person}{Gang Huang}, {and} \bibinfo{person}{Xuanzhe Liu}.} \bibinfo{year}{2021}\natexlab{}.
\newblock \showarticletitle{TaintStream: fine-grained taint tracking for big data platforms through dynamic code translation} \emph{(\bibinfo{series}{ESEC/FSE 2021})}. \bibinfo{publisher}{Association for Computing Machinery}, \bibinfo{address}{New York, NY, USA}, \bibinfo{pages}{806–817}.
\newblock
\showISBNx{9781450385626}
\urldef\tempurl%
\url{https://doi.org/10.1145/3468264.3468532}
\showDOI{\tempurl}


\bibitem[Yao et~al\mbox{.}(2023)]%
        {yao2023react}
\bibfield{author}{\bibinfo{person}{Shunyu Yao}, \bibinfo{person}{Jeffrey Zhao}, \bibinfo{person}{Dian Yu}, \bibinfo{person}{Nan Du}, \bibinfo{person}{Izhak Shafran}, \bibinfo{person}{Karthik Narasimhan}, {and} \bibinfo{person}{Yuan Cao}.} \bibinfo{year}{2023}\natexlab{}.
\newblock \showarticletitle{{ReAct}: Synergizing Reasoning and Acting in Language Models}. In \bibinfo{booktitle}{\emph{International Conference on Learning Representations (ICLR)}}.
\newblock


\bibitem[Yuan et~al\mbox{.}(2024)]%
        {foundation_model_as_firmware}
\bibfield{author}{\bibinfo{person}{Jinliang Yuan}, \bibinfo{person}{Chen Yang}, \bibinfo{person}{Dongqi Cai}, \bibinfo{person}{Shihe Wang}, \bibinfo{person}{Xin Yuan}, \bibinfo{person}{Zeling Zhang}, \bibinfo{person}{Xiang Li}, \bibinfo{person}{Dingge Zhang}, \bibinfo{person}{Hanzi Mei}, \bibinfo{person}{Xianqing Jia}, \bibinfo{person}{Shangguang Wang}, {and} \bibinfo{person}{Mengwei Xu}.} \bibinfo{year}{2024}\natexlab{}.
\newblock \showarticletitle{Mobile Foundation Model as Firmware}. In \bibinfo{booktitle}{\emph{Proceedings of the 30th Annual International Conference on Mobile Computing and Networking}} (Washington D.C., DC, USA) \emph{(\bibinfo{series}{ACM MobiCom '24})}. \bibinfo{publisher}{Association for Computing Machinery}, \bibinfo{address}{New York, NY, USA}, \bibinfo{pages}{279–295}.
\newblock
\showISBNx{9798400704895}
\urldef\tempurl%
\url{https://doi.org/10.1145/3636534.3649361}
\showDOI{\tempurl}


\bibitem[Zheng et~al\mbox{.}(2023)]%
        {zheng2023codegeex}
\bibfield{author}{\bibinfo{person}{Qinkai Zheng}, \bibinfo{person}{Xiao Xia}, \bibinfo{person}{Xu Zou}, \bibinfo{person}{Yuxiao Dong}, \bibinfo{person}{Shan Wang}, \bibinfo{person}{Yufei Xue}, \bibinfo{person}{Zihan Wang}, \bibinfo{person}{Lei Shen}, \bibinfo{person}{Andi Wang}, \bibinfo{person}{Yang Li}, \bibinfo{person}{Teng Su}, \bibinfo{person}{Zhilin Yang}, {and} \bibinfo{person}{Jie Tang}.} \bibinfo{year}{2023}\natexlab{}.
\newblock \showarticletitle{CodeGeeX: A Pre-Trained Model for Code Generation with Multilingual Benchmarking on HumanEval-X}. In \bibinfo{booktitle}{\emph{Proceedings of the 29th ACM SIGKDD Conference on Knowledge Discovery and Data Mining}}. \bibinfo{pages}{5673--5684}.
\newblock


\bibitem[Zhuo et~al\mbox{.}(2024)]%
        {zhuo2024bigcodebench}
\bibfield{author}{\bibinfo{person}{Terry~Yue Zhuo}, \bibinfo{person}{Minh~Chien Vu}, \bibinfo{person}{Jenny Chim}, \bibinfo{person}{Han Hu}, \bibinfo{person}{Wenhao Yu}, \bibinfo{person}{Ratnadira Widyasari}, \bibinfo{person}{Imam Nur~Bani Yusuf}, \bibinfo{person}{Haolan Zhan}, \bibinfo{person}{Junda He}, \bibinfo{person}{Indraneil Paul}, {et~al\mbox{.}}} \bibinfo{year}{2024}\natexlab{}.
\newblock \showarticletitle{BigCodeBench: Benchmarking Code Generation with Diverse Function Calls and Complex Instructions}.
\newblock \bibinfo{journal}{\emph{arXiv preprint arXiv:2406.15877}} (\bibinfo{year}{2024}).
\newblock


\end{thebibliography}

\end{document}